\documentclass{article}

\PassOptionsToPackage{numbers, compress}{natbib}
\usepackage{algcompatible}
\usepackage[final]{neurips_2023}

\usepackage[utf8]{inputenc} % allow utf-8 input
\usepackage[T1]{fontenc}    % use 8-bit T1 fonts
\usepackage{amsmath}
\usepackage{amsfonts}       % blackboard math symbols
\usepackage{url}            % simple URL typesetting
\usepackage{booktabs}       % professional-quality tables
\usepackage{nicefrac}       % compact symbols for 1/2, etc.
\usepackage{microtype}      % microtypography
\usepackage{xcolor}         % colors
\usepackage{bm}
\usepackage{diagbox}
\usepackage{oplotsymbl}
\usepackage{pgfplots}
\usepackage{wrapfig}
\usetikzlibrary{patterns}
\usepackage{subcaption}

\usepackage[ruled,vlined]{algorithm2e}

\SetKwInput{KwInput}{Input}                % Set the Input
\SetKwInput{KwOutput}{Output}              % set the Output

\DeclareRobustCommand\encircle[1]{\tikz[baseline=(char.base)]{\node[shape=circle,fill,inner sep=1pt] (char) {\textcolor{white}{#1}}}}

\newif\ifdraft
\drafttrue
\usepackage{xparse}
\usepackage{xspace}
\usepackage{algorithm}% http://ctan.org/pkg/algorithm%
\usepackage{algpseudocode}% http://ctan.org/pkg/algorithmicx
\usepackage{fixltx2e}
\usepackage{tikz}
\usepackage{float}
\usepackage{multirow}
\usepackage{multicol}
\usepackage{colortbl}
\usepackage{array}
\newcommand{\PreserveBackslash}[1]{\let\temp=\\#1\let\\=\temp}
\newcolumntype{C}[1]{>{\PreserveBackslash\centering}p{#1}}
\newcolumntype{R}[1]{>{\PreserveBackslash\raggedleft}p{#1}}
\newcolumntype{L}[1]{>{\PreserveBackslash\raggedright}p{#1}}

\usepackage{microtype}
\usepackage{graphicx}
\usepackage{mathtools}
\usepackage{float}
\usepackage{subcaption}
\usepackage{booktabs} % for professional tables
\usepackage[shortlabels]{enumitem}

\usepackage{amssymb}% http://ctan.org/pkg/amssymb
\usepackage{pifont}% http://ctan.org/pkg/pifont

\usepackage{bbold}

\usepackage{hyperref,amssymb,enumitem}

\usepackage{tabu}
\usepackage{xcolor}

\setlist[itemize]{leftmargin=*}
\setlist[enumerate]{leftmargin=*}

\newcommand*{\rej}{{\ooalign{\lower.3ex\hbox{$\sqcup$}\cr\raise.4ex\hbox{$\sqcap$}}}}

\usepackage{stackengine}

\usepackage{xcolor}

\renewcommand{\algorithmicrequire}{\textbf{Input: }}
\renewcommand{\algorithmicensure}{\textbf{Output: }}

\newcommand{\ie}{\textit{i.e.,}\@\xspace}
\newcommand{\eg}{\textit{e.g.,}\@\xspace}

\DeclareRobustCommand\encircle[1]{\tikz[baseline=(char.base)]{\node[shape=circle,fill,inner sep=1pt] (char) {\textcolor{white}{#1}}}}

\graphicspath{{images/}}

\newcommand{\name}{B4B\@\xspace}

\newcommand{\Legit}{Legit\@\xspace}
\newcommand{\Attacker}{Attack\@\xspace}
\newcommand{\Sybil}{Sybil\@\xspace}
\newcommand{\efra}{\mathcal{E}_f}

\newcommand{\enc}{f_v}

\usepackage{booktabs,arydshln}
\makeatletter
\def\adl@drawiv#1#2#3{%
        \hskip.5\tabcolsep
        \xleaders#3{#2.5\@tempdimb #1{1}#2.5\@tempdimb}%
                #2\z@ plus1fil minus1fil\relax
        \hskip.5\tabcolsep}
\newcommand{\cdashlinelr}[1]{%
  \noalign{\vskip\aboverulesep
           \global\let\@dashdrawstore\adl@draw
           \global\let\adl@draw\adl@drawiv}
  \cdashline{#1}
  \noalign{\global\let\adl@draw\@dashdrawstore
           \vskip\belowrulesep}}
\makeatother
\usepackage{cleveref}
\crefalias{prop}{proposition} % <-- new

\newcommand{\nlp}[1]{}
\newcolumntype{x}[1]{>{\centering\arraybackslash\hspace{0pt}}p{#1}}
\ifdraft
\usepackage[colorinlistoftodos,textsize=scriptsize]{todonotes}
\newcommand{\mynote}[1]{\textcolor{red}{[note: #1]}}

\newcommand{\mytodo}[1]{\textcolor{red}{[todo: #1]}}
\newcommand{\mycomment}[1]{\textcolor{red}{[comment: #1]}}
\newcommand{\chris}[1]{\textcolor{red}{Chris: #1}}
\newcommand{\ahmad}[1]{\textcolor{darkpastelgreen}{[Ahmad: #1]}}
\newcommand{\vinith}[1]{\textcolor{blue}{Vinith: #1}}
\newcommand{\yunxiang}[1]{\textcolor{cyan}{Yunxiang: #1}}
\newcommand{\xiao}[1]{\textcolor{blue}{xiao: #1}}
\definecolor{chocolate(traditional)}{rgb}{0.48, 0.25, 0.0}

\definecolor{darkpastelgreen}{rgb}{0.01, 0.75, 0.24}
\newcommand{\natalie}[1]{\textcolor{darkpastelgreen}{natalie: #1}}
\definecolor{pistachio}{rgb}{0.58, 0.77, 0.45}

\newcommand{\jonas}[1]{\textcolor{violet}{[Jonas: #1]}}

\newcommand{\adelin}[1]{\textcolor{red}{[Adelin: #1]}}
\newcommand{\mohammad}[1]{\textcolor{red}{[Mohammad: #1]}}

\definecolor{amber(sae/ece)}{rgb}{1.0, 0.49, 0.0}
\newcommand\adam[1]{{\textcolor{red}{[Adam: #1]}}}
\newcommand{\sierra}[1]{\textcolor{blue}{[Sierra: #1]}}
\newcommand{\armin}[1]{\textcolor{cyan}{[Armin: #1]}}
\newcommand{\nikita}[1]{\textcolor{cyan}{[Nikita: #1]}}
\newcommand{\franzi}[1]{\textcolor{purple}{[Franzi: #1]}}

\else

\usepackage[disable]{todonotes}
\newcommand{\chris}[1]{}
\newcommand{\franzi}[1]{}
\newcommand{\vinith}[1]{}
\newcommand{\adam}[1]{}
\newcommand{\yunxiang}[1]{}
\newcommand{\natalie}[1]{}
\newcommand{\jonas}[1]{}
\newcommand{\adelin}[1]{}
\newcommand{\mynote}[1]{}
\newcommand{\xiao}[1]{}
\newcommand{\mytodo}[1]{}
\newcommand{\mycomment}[1]{}
\newcommand{\ahmad}[1]{}
\newcommand{\mohammad}[1]{}
\newcommand{\sierra}[1]{}
\newcommand{\armin}[1]{}
\newcommand{\nikita}[1]{}

\fi

\usepackage{amsmath,amsfonts,bm}

\def\eqref#1{equation~\ref{#1}}
\def\1{\bm{1}}

\DeclareMathAlphabet{\mathsfit}{\encodingdefault}{\sfdefault}{m}{sl}
\SetMathAlphabet{\mathsfit}{bold}{\encodingdefault}{\sfdefault}{bx}{n}

\usepackage{hyperref}       % hyperlinks
\newcommand{\ourtitle}{Bucks for Buckets (B4B): Active Defenses Against Stealing Encoders}
\title{\ourtitle}

\author{
 Jan Dubi{\'n}ski $^{1,2}$\thanks{Corresponding authors: jan.dubinski.dokt@pw.edu.pl and adam.dziedzic@cispa.de}\ \ \thanks{Equal contribution.} %\orcidID{0000-0002-2568-0132}
 \quad
 Stanis{\l{}}aw Pawlak  $^{1\dagger}$%\orcidID{0000-0001-7511-9995}
 \quad
 Franziska Boenisch $^{4\dagger}$\\
 \textbf{Tomasz Trzci{\'n}ski $^{1,2,3}$}%\orcidID{0000-0002-1486-8906}
 \quad
 \textbf{Adam Dziedzic $^{4*}$\thanks{Project Lead.} }\\
$^1$\small Warsaw University of Technology \quad $^2$\small IDEAS NCBR \quad $^3$\small Tooploox\\
$^4$\small CISPA Helmholtz Center for Information Security \quad 
}

\begin{document}

\maketitle
\vspace{-1em}
\begin{abstract}
\vspace{-1em}

%v3
%v2
Machine Learning as a Service (MLaaS) APIs provide ready-to-use and high-utility encoders that generate vector representations for given inputs. Since these encoders are very costly to train, they become lucrative targets for model stealing attacks during which an adversary leverages query access to the API to replicate the encoder locally at a fraction of the original training costs. We propose \textit{Bucks for Buckets (B4B)}, the first \textit{active defense} that prevents stealing while the attack is happening without degrading representation quality for legitimate API users. Our defense relies on the observation that the representations returned to adversaries who try to steal the encoder's functionality cover a significantly larger fraction of the embedding space than representations of legitimate users who utilize the encoder to solve a particular downstream task.
B4B leverages this to adaptively adjust the utility of the returned representations according to a user's coverage of the embedding space. To prevent adaptive adversaries from eluding our defense by simply creating multiple user accounts (sybils), B4B also individually transforms each user's representations. This prevents the adversary from directly aggregating representations over multiple accounts to create their stolen encoder copy. Our active defense opens a new path towards securely sharing and democratizing encoders over public APIs.

\end{abstract}

\section{Introduction}

%\adam{The text for this content should be moved from this google doc: \url{https://docs.google.com/document/d/15ExbuJ-raRVIF5MIUOMJ7t_oFPWzt3gRbsgMBlc8O1s/edit?usp=sharing}.}
In model stealing attacks, adversaries extract a machine learning model exposed via a public API by repeatedly querying it and updating their own stolen copy based on the obtained responses. 
Model stealing was shown to be one of the main threats to the security of machine learning models in practice~\citep{MicrosoftSurvey}. %\tf{this sentence still sounds empty and not well connected to the next. Maybe we can extend it.}
%[]\tf{@Adam, please cite the Microsoty survey here}. \tf{maybe make clear that stealing and extraction means the same thing}
Also in research, since the introduction of the first extraction attack against classifiers~\citep{tramer2016usenix}, a lot of work on improving stealing~\citep{kariyappa2021maze,orekondy2019knockoff,tramer2016usenix,DataFreeExtract}, extending it to different model types~\cite{carlini2021extracting,shen2022model}, and proposing adequate defenses~\citep{powDefense,jia2021entangled,juuti2019prada,digital_wm} has been put forward.
With the recent shift in learning paradigms from supervised to self supervised learning (SSL), especially the need for new defenses becomes increasingly pressing.
From an academic viewpoint, the urge arises because it was shown that SSL models (\textit{encoders}) are even more vulnerable to model stealing~\citep{SSLextraction,StolenEncoder,ContSteal} than their supervised counterparts.
This is because whereas supervised models' output is low dimensional, \eg per-class probabilities or pure labels, SSL encoders output high-dimensional representation vectors that encode a larger amount of information and thereby facilitate stealing.
In addition, from a practical industry's viewpoint, defenses are required since many popular API providers, such as Cohere, OpenAI, or Clarify~\citep{Clarifai,Cohere,OpenAI} already expose their high-value SSL encoders via APIs to a broad range of users.

% Model stealing attacks and defenses have become the new area of great interest from both research and industry viewpoints. 
% Since the dawn of the first work on model extraction~\citep{tramer2016usenix}, there has been an increasing effort to assess the threat of model stealing~\citep{jagielski2020fidelity} and a new spectrum of defenses~\citep{powDefense,juuti2019prada}. The new emerging paradigm of SSL encoders shifted the focus to the emerging APIs that return representations for given inputs instead of low-dimensional outputs~\citep{Clarifai,Cohere,OpenAI}. Recent work showed that SSL encoders are even more vulnerable to model stealing~\citep{SSLextraction,StolenEncoder,ContSteal} than previous supervised models. 

Most of the current defenses against encoder stealing are \textit{reactive}, \ie they do not actively prevent the stealing but rather aim at detecting it by adding watermarks to the encoder~\citep{SSLguard, SSLextraction} or performing dataset inference to identify stolen copies~\citep{SSLdatasetinference}.
Since at the point of detection, the damage of stealing has already been inflicted,
%and it is unclear how ownership resolution of encoders can be settled in practice 
we argue that reactive defenses intervene too late and we advocate for \textit{active} defenses that prevent stealing while it is happening.
Yet, active defenses are challenging to implement because they not only need to prevent stealing but also should preserve the utility of representations for legitimate users.
The only existing active defense against encoder stealing~\citep{StolenEncoder} falls short on this latter aspect since it significantly degrades the quality of representations for all users.

\begin{figure}[t]
\vspace{-0.cm}
    \centering
    \includegraphics[width=\linewidth]{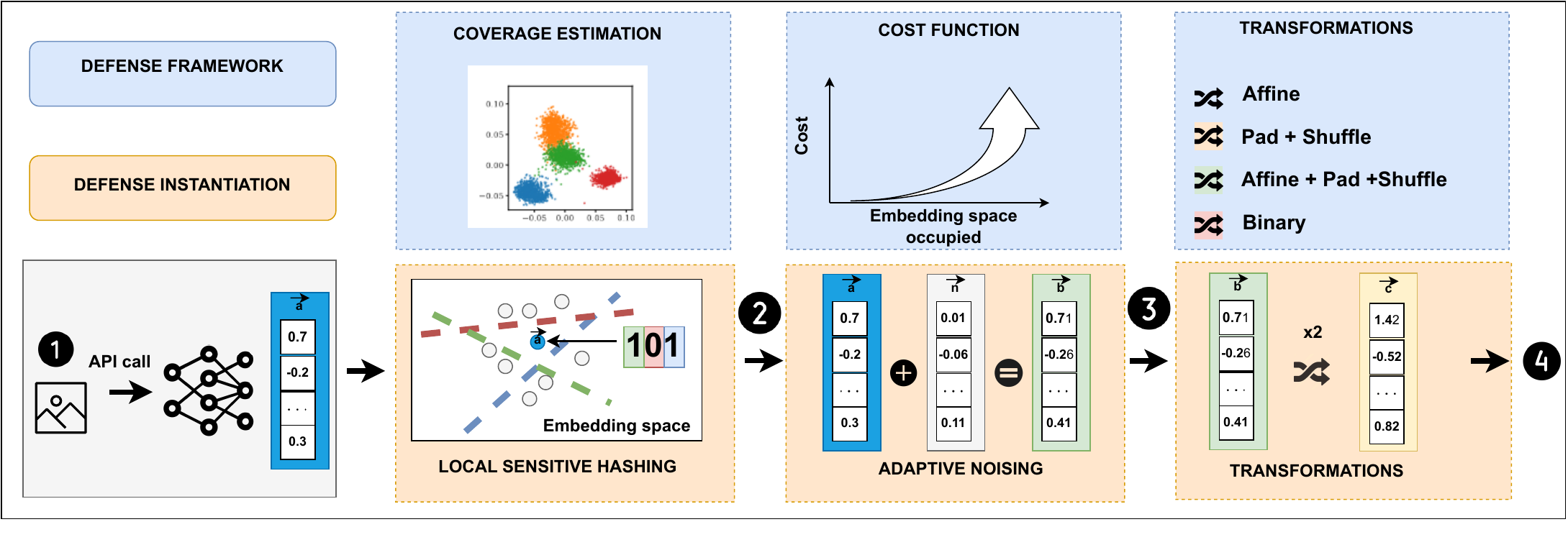} %width=\linewidth
    %\caption{\textbf{Overview of P2P Learning}. Every peer can act as a client (to compute updates) or server (to aggregate updates). \textcolor{blue}{Blue} represents semi-honest while \textcolor{red}{red} is the malicious setting.}
    \caption{
    \textbf{Overview of \name.} In the upper part, we present our B4B framework that consists of three modular building blocks: (1) A coverage estimation to track the fraction of embedding space covered by the representations returned to each user, (2) a cost function that serves to map the coverage to a concrete penalty to prevent stealing, and (3) per-user transformations that are applied to the returned representations to prevent sybil attacks. 
    In the lower part, we present a concrete instantiation of B4B and the operation flow of our defense: \encircle{1} The API calculates representations for the incoming queries. \encircle{2} We instantiate the coverage estimation with local sensitive hashing and estimate the covered space as the fraction of \textit{hash buckets} occupied.  We calibrate the costs by adding noise to the representations according to the coverage.
    %For a single-account defense, we compute the query cost based on the number of buckets filled in by the querying user. The output representations are noised based on the computed cost to lower the fidelity of the extraction. 
    \encircle{3} We apply a set of transformations on a per-user basis. \encircle{4} The noised and transformed representations are returned to the user.
    }
    \label{fig:b2b-overview}
\vspace{-0.cm}
\end{figure}

To close the gap between required and existing defenses, we propose \textit{Bucks for Buckets (B4B)}, the first active defense against encoder stealing that does not harm utility for legitimate users.
B4B leverages the observation that the representations returned to adversaries who try to steal the encoder's functionality cover a significantly larger fraction of the full embedding space than representations of legitimate users who utilize the encoder to solve a particular downstream task.
To turn this observation into a practical defense, B4B is equipped with three modular building blocks:
(1) The first building block is a tracking mechanism that continuously estimates the fraction of the embedding space covered by the representations returned to each user.
The intuition why this is relevant is that by covering large fractions of the embedding space, the representations will suffice for an adversary to reproduce the encoder's functionality, \ie to successfully steal it.
(2) B4B's second building block consists of a cost function to translate the covered fraction of the embedding space into a concrete penalty.
We require this cost function to significantly penalize adversaries trying to steal the model while having only a minimal effect on legitimate users.
(3) The third building block contains transformations that can be applied to the representations on a per-user basis to prevent adaptive attackers from circumventing our defense by creating multiple user accounts (sybils) and distributing their queries over these accounts such that they minimize the overall cost.
We present the different building blocks of B4B in \Cref{fig:b2b-overview}.
% \franzi{B4B is a framework that consists of 3 building blocks:
% 1. Tracking embedding space coverage for user representations
% 2. The cost function that based on the coverage noises the user's representations
% 3. the transformations that fight off sybils
% All of them can be instantiated with different functions.
% We propose a concrete instantiation based on...}

While B4B's modularity enables different instantiations of the three building blocks, we propose a concrete end-to-end instantiation to showcase the practicability of our approach.
To implement tracking of the covered embedding space, we employ \textit{local sensitive hashing} that maps any representation returned to a given user into a set of hash \textbf{\textit{buckets}}.
%Costs for incoming user queries are then adapted according to the fraction of buckets already occupied by their representations.
%Our B4B then adaptively adjusts the \textit{costs} for incoming user-queries according to the fraction of buckets already occupied by their representations.
We base our cost function (\ie the \textit{\textbf{"bucks"}}) on utility and make B4B add noise to the representations with a magnitude that increases with the number of buckets occupied by the given user. % to the representations before returning them. 
%\footnote{Note that our B4B allows to integrate multiple alternative cost functions, that rely on, for example, monetary costs, \ie increase the per-query-batch price with increasing coverage, or on costs in terms of an increasing amount of computations required by the users to obtain their representations, similar to a Proof-of-Work~\citep{powDefense}.}
%We design a cost function for B4B that adds noise with a magnitude that increases with the number of buckets occupied by a given user to their representations before returning them. 
While the scale of noise added to legitimate users' representations does not harm their downstream performance due to their small embedding space coverage, the representations returned to an adversary become increasingly noisy---significantly degrading the performance of their stolen encoder.
Finally, we rely on a set of transformations (\eg affine transformations, shuffling, padding) that preserve downstream utility~\citep{SSLdatasetinference}. 
While, as a consequence, legitimate users remain unaffected by these transformations, 
% To prevent adaptive attackers from circumventing our defense by creating multiple user-accounts (sybils) and distributing their queries over these accounts such that they minimize each account's utility drop, we equip B4B with per-user transformations that are applied to their returned representations.
% Prior work has observed that certain types of transformations (\eg affine transformations, shuffling, padding) do not harm the performance of downstream tasks~\citep{SSLdatasetinference} and hence preserve the utility for legitimate users within B4B.
% %As a result, B4B's per-user transforming do not harm utility for legitimate users.
% In contrast, an 
adversaries cannot directly combine the representations obtained through different sybil accounts anymore to train their stolen copy of the encoder.
Instead, they first have to remap all representations into the same embedding space, which we show causes both query and computation overhead and still reduces the performance of the stolen encoder. %\tf{These latter, we need to verify them in the experiments and potentially weaken the text.}

In summary, we make the following contributions:
\begin{enumerate}
    \item We present \name, the first active defense against encoder stealing that does not harm legitimate users' downstream performance. \name's three building blocks enable penalizing adversaries whose returned representations cover large fractions of the embedding space and prevent sybil attacks.
    \item We propose a concrete instantiation of \name that relies on local sensitive hashing and decreases the quality of representations returned to a user once their representations fill too many hash buckets. %We show how to calibrate the cost function in B4B and 
    %To prevent Sybil attackers, who create many fake accounts, we output differently transformed representations to each user such that the remapping to a unified space between representations is prohibitively costly.
    %Therefore, we rely on per-user transformations for the returned representations which makes combining representations from different accounts inefficient for encoder stealing.
    \item We provide an end-to-end evaluation of our defense to highlight its effectiveness in offering high utility representations for legitimate users and degrading the performance of stolen encoders in both the single and the sybil-accounts setup.
    %that shows how to calibrate the defense against a single-user attacker, how the transformations per account effectively prevent Sybil adversaries, and finally present an end-to-end experiment which shows how our defense performs in realistic scenarios.
\end{enumerate}
\section{Related Work}
%\vspace{-1em}
\label{sec:related-work}

%\franzi{@Adam, where do these paragraphs come from? Just such that I can insert the right bibliography items.}
%\adam{This is the first draft on the related work.}
\textbf{Model Extraction Attacks.}
\label{sec:extaction-attacs}
The goal of the model extraction attacks is to replicate the functionality of a victim model $f_v$ trained on a dataset $D_v$. An attacker has a black box access to the victim model and uses a stealing dataset $D_s = \{q_i, f_v(q_i)\}_{i=1}^n$, consisting of queries $q_i$ and the corresponding outputs $f_v(q_i)$ returned by the victim model, to train a stolen model $f_s$. 
% The most common attacks target the fidelity or accuracy of the extracted model. 
Model extraction attacks have been shown against various types of models including classifiers~\citep{jagielski2020fidelity,tramer2016usenix} and encoders~\citep{SSLextraction,ContSteal}. 

\textbf{Self Supervised Learning and Encoders.}
\label{sec:background:ssl}
%\franzi{Here, we need background on all the frameworks we use and a little on how SSL is working.}
%Self-supervised learning (SSL) 
SSL is an increasingly popular machine learning paradigm. It trains encoder models to generate representations from complex inputs without relying on explicit labels. These representations encode useful features of a given input, enabling efficient learning for multiple downstream tasks. 
Many SSL frameworks have been proposed~\citep{dino_2021_ICCV,simclr,SimSiam,byol,he2022masked,zbontar2021barlow}. In our work, we focus on the two popular SSL vision encoders, namely SimSiam~\citep{SimSiam} and DINO~\citep{dino_2021_ICCV}, which return high-quality representations that achieve state-of-the-art performance on downstream tasks when assessed by training a linear classifier directly on representations. %\tf{@Adam, without knowing the papers, I have no chance on understanding the next 3 sentences. Let's rewrite together. Adam: I rewrote that.} 
SimSiam trains with two Siamese encoders with directly shared weights. A prediction MLP head is applied to one of the encoders $f_1$, and the other encoder $f_2$ has a stop-gradient, where both operations are used for avoiding collapsing solutions. In contrast, DINO shares only architecture (not weights) between a student $f_1$ and a teacher model $f_2$, also with the stop-gradient operation, but not the prediction head. While SimSiam uses convolutional neural networks (CNNs), DINO also employs vision transformers (ViTs). Both frameworks use a symmetrized loss of the form $\frac{1}{2}g(f_1(x_1),f_2(x_2)) + \frac{1}{2}g(f_1(x_2),f_2(x_1))$ in their optimization objectives, where $g(\cdot,\cdot)$ is negative cosine similarity  for SimSiam and cross-entropy for DINO. SimSiam and DINO's similarities and differences demonstrate our method's broad applicability across SSL frameworks. More details can be found in \Cref{app:related-work}.

\textbf{Stealing Encoders.}
\label{sec:background:vit}
%\franzi{I took the freedom to write in the following section everything I don't understand. As a non-expert in the area, I might be a possible reviewer and I think that things should be self-explainatory exactly for this audience.}
%Vision transformers~\citep{dosovitskiy2020image, touvron2021training, park2022vision}, inspired by transformer models from the NLP domain, have been used successfully in the vision domain.
%Their performance has, in some cases, exceeded convolutional neural networks \franzi{This statement seems vague. And in what terms is performance measured here? Downstream tasks, I assume? And which CNNs are they compared against}. 
% In our work, we use vision encoders~\citep{dino_2021_ICCV,SimSiam}, which return high-quality representations that achieve SoTA performance on downstream tasks when assessed by training a linear classifier directly on representations. 
%DINO trains student and %(momentum) \franzi{what does momentum mean here?} - this is the other way to say that teacher's parameters are updated with an (exponential moving) average of the student's parameters.
%teacher encoders, both with the same architecture but different parameters, where the teacher is updated with an (exponential moving) average of the student.
%\textbf{Stealing Encoders.}
The stealing of SSL encoders was shown to be extremely effective~\citep{SSLextraction,StolenEncoder,ContSteal}. 
The goal of extracting encoders is to maximize the similarity of the outputs from the stolen local copy and the original representations output by the victim encoder.
Therefore, while training the stolen copy, the adversary either imitates a self-supervised training using a contrastive loss function, \eg InfoNCE~\cite{simclr} or SoftNN~\cite{softnn-frosst19a} or directly matches both models' representations via the Mean Squared Error (MSE) loss. To reduce the number of queries sent to the victim encoder, the attack proposed in~\citep{StolenEncoder} leverages the key observation that the victim encoder returns similar representations for any image and its augmented versions. Therefore, a given image can be sent to the victim while the stolen copy is trained using many augmentations of this image, where the representation of a given augmented image is approximated as the one of the original image produced by the victim encoder.

\textbf{Defending Encoders.}
Recently, watermarking~\citep{adi2018turning,jia2021entangled,uchida2017embedding} methods have been proposed to detect stolen encoders~\citep{SSLguard,SSLextraction,Wu2022WatermarkingPE}. Many of these approaches use downstream tasks to check if a watermark embedded into a victim encoder is present in a suspect encoder.
Dataset inference~\citep{maini2021dataset} is another type of encoder ownership resolution.
%It enables a model owner or a third-party arbitrator to attribute the ownership of a potentially stolen model.
It uses the victim's training dataset as a unique signature, leveraging the following observation: 
for a victim encoder trained on its private data as well as for its stolen copies, the distribution of the representations generated from the victim's training data differs from the distribution of the representations generated on the test data.
In contrast, for an independently trained encoder, these two distributions cannot be distinguished, allowing the detection of stolen copies~\citep{SSLdatasetinference}.
%For modeling the distributions, Gaussian Mixture Models (GMMs) are trained on a fraction of the private training data and applied to a disjoint fraction of the training data and the test data. An encoder is identified as a victim or stolen copy if the \prob on private representations is significantly higher than on the test representations.
However, all the previous methods are \textit{reactive} and aim at detecting the stolen encoder instead of \textit{actively} preventing the attack. 
The only preliminary active defenses for encoders were proposed by~\cite{SSLextraction,StolenEncoder}. They either perturb or truncate the answers to poison the training objective of an attacker. These operations were shown to harm substantially the performance of legitimate users, which renders the defense impractical. In contrast, our \name has negligible impact on the quality of representations returned to legitimate users. %and thus, being not usable.

\section{Actively Defending against Model Stealing with B4B}
\label{sec:method}
B4B aims at actively preventing model stealing while preserving high-utility representations for legitimate users. 
Before introducing the three main building blocks of B4B, namely (1) the estimation of embedding space coverage, (2) the cost function, and (3) the transformation of representations (see \Cref{fig:b2b-overview}), we detail our threat model and the observation on embedding space coverage that represents the intuition behind our approach.

\subsection{Threat Model and Intuition}

%\adam{@Franzi: this is what we should also include probably here. Our defense targets real-world scenario where the SSL service provider exposes encoder trained on a large amount of unlabeled data to enable effective extraction of features for diverse downstream tasks. 
%The standard users who can benefit from the API are those who issue queries to handle their specific downstream tasks. On the other hand, adversaries would use much more diverse training set to steal the whole representations space. 
%For instance, if the encoder behind the API is trained on ImageNet dataset, then the standard users are expected to query it for downstream tasks, such as CIFAR10 or SVHN. On the other hand, if the encoder is trained on CIFAR10, then the expected downstream tasks are MNIST or Fashion MNIST. In general, the dataset used to train the encoder should be much richer and expressive than a single downstream task. 
%
%}

%\paragraph{Threat Model.}
Our setup and the resulting threat model are inspired by public APIs, such as Cohere, OpenAI, or Clarify~\citep{Clarifai,Cohere,OpenAI} that expose encoders to users through a pre-defined interface. 
These encoders are trained using SSL on large amounts of unlabeled data, often crawled from the internet, and therefore from diverse distributions.
We notice that to provide rich representations to multiple users, the training dataset of the encoder needs to be significantly more diverse than the individual downstream tasks that the users query for representations.
For instance, if the encoder behind the API is trained on the ImageNet dataset, then the legitimate users are expected to query the API for downstream tasks, such as CIFAR10 or SVHN.
Similarly, if the encoder is trained on CIFAR10, the expected downstream tasks are MNIST or Fashion MNIST.
Yet, in the design of our defense, we consider adversaries who can query the encoder with arbitrary inputs to obtain high-dimensional representation vectors from the encoder.
Our defense is independent of the protected encoder's architecture and does not rely on any assumption about the adversary's data and query strategy. %\tf{Can we say it like this? Adam: I think so.}

\definecolor{mygreen2}{RGB}{0, 180, 18}
\begin{wrapfigure}{r}{5cm}
%\begin{figure}[h!]
\vspace{-0.0cm}
    \centering
    \includegraphics[width=0.3\textwidth, trim={1cm 0.8cm 1cm 2cm},clip]{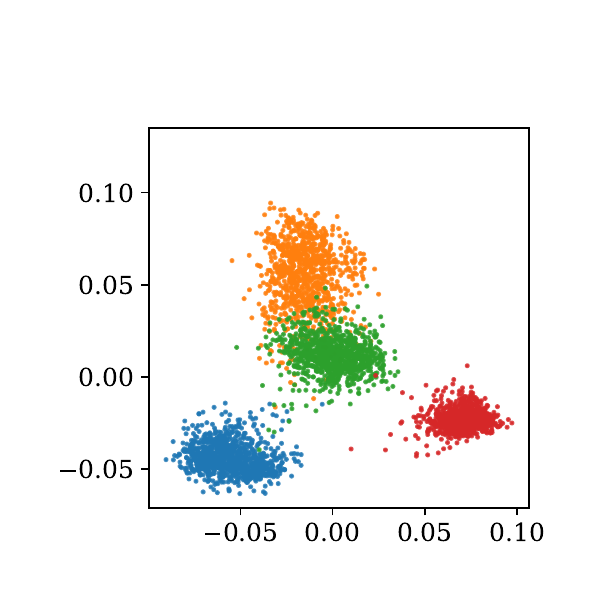} 
    %\includegraphics[width=0.25\textwidth, trim={1cm 1cm 1cm 1cm}]{figures/clusters/full_repr_dataset_clusters_stl10_4x4.pdf} 
    %\vspace{-2ex}
    %\caption{\textbf{Lower Dimension Representation of Different Representations.} We map the representations obtained for different downstream tasks to a two-dimensional space. We observe that the different downstream tasks form clusters.}
    \caption{\textbf{Representations from Different Tasks Occupy Different Sub-Spaces of the Embedding Space. Presented for {\color{blue}FashionMNIST}, {\color{orange}SVHN}, {\color{mygreen2}CIFAR10}, and {\color{red}STL10}.}
    %\janek{Please choose pca from full representation (encoder hidden state) - bottom, or last 4 layers - top}
    %\adam{The legend has to be changed. We should have simply: FashionMNIST, STL10, SVHN, and CIFAR10. If we add ImageNet, it would probably cover more of the representation space - this might be shown in a separate picture.}
    %\franzi{I need to color the names}
    %\ta{Why do we have this figure instead of the one with the very small space occupied for FashionMNIST?}
    }
    \vspace{-0.3cm}
    \label{fig:clusters}
%\end{figure}
\end{wrapfigure}

%\franzi{The following might still read a little repetitive. I will reiterate} 
We argue that even in this restricted setup, our defense can distinguish between adversaries and legitimate users by analyzing the distribution of representations returned to them.
%\textbf{Adversaries can be distinguished from legitimate users through the distribution of their representations in the embedding space.}
%We build this argument on the observation that the
In \Cref{fig:clusters}, by using PCA to project representations for different datasets to a two-dimensional space, we visualize that  representations for different downstream tasks cluster in \textit{disjoint} and \textit{small sub-spaces} of the full embedding space.  
The representations were obtained from a SimSiam encoder originally trained on ImageNet (we observe similar clustering for DINO shown in \Cref{app:epxeriments}). %\tf{@Adam, @Janek: verify is this is correct. Adam: this is correct. If time allows, we can also show it on SimSiam, but this can be done later for the appendix submission.}
As a result, legitimate users can be characterized by their representations' small coverage of the embedding space.
In contrast, the adversary does not aim at solving a particular downstream task. 
They instead would want to obtain representations that cover large fractions of the embedding space.
This enables reproducing the overall functionality of the encoder (instead of only learning some local task-specific behavior).
%An adversary, to steal the encoder, therefore, needs to query with a versatile dataset containing multiple distributions whose returned representations cover larger fraction of the embedding space than the representations of a single task.
%The intuition why an adversary would want to obtain representations that cover large fractions of the embedding space is that this enables to reproduce the overall functionality of the encoder (instead of only learning some local behavior).
Indeed, it has been empirically shown by prior work, such as \citep{SSLextraction}, that stealing with multiple distributions, \eg by relying on the complex ImageNet dataset, yields higher performance of the stolen encoder on various downstream tasks than stealing with a downstream dataset, such as CIFAR10.
%From prior work~\citep{SSLextraction} we know that in order to steal high-utility encoders that will achieve high performance on various downstream tasks, 
%it is not sufficient to steal with one single downstream dataset. % imagenet infonce is usually 3-5 % better than stealing with CIFAR10 on all other downstream tasks.
%Instead, the more varied the stealing dataset is (for example ImageNet), the better the performance.
%one requires a versatile query dataset with multiple distributions. % that cover large fractions of the embedding space.
%With the observation from \Cref{fig:clusters}, we understand that the multiple data distributions lead to the representations covering larger fraction of the space.
%By covering large fractions of the embedding space, the representations will suffice for the user to reproduce the overall functionality of the encoder.
%Combining this knowledge with our observation naturally equips us with the ability to identify and penalize adversaries who want to steal the encoder:
As a result, intuitively, we can identify and penalize adversaries based on their coverage of the embedding space, which will be significantly larger than the coverage of legitimate users.
%For an optimal fine-tuning of our defense, it is, therefore, helpful, if the API provider formulates an expectation on the complexity and diversity of the downstream tasks (in contrast to the diversity of the training data).
%The final tuning of the defense to a specific encoder exposed via the API should define the scope of the expected target downstream tasks, for instance, in terms of their occupation of the embedding space.
%While the representations for legitimate users cover small areas of the entire embedding space, adversaries' representations cover larger areas.
%Intuitively, to identify and penalize adversaries, we can, thus, rely on the coverage of embedding space.
We leverage this intuition to build our B4B defense and present our three main building blocks in the following sections.

\subsection{Building Block 1: Coverage Estimation of the Embedding Space}
\label{sub:block1coverage}

The first building block of our B4B serves to estimate and continuously keep track of the fraction of the embedding space occupied by any given user.
Let $\mathcal{E}$ denote our embedding space of dimension~$s$, further, let $U$ be a user with a query dataset $D={q_1,\dots,q_n} \in \mathcal{D}$ and let $\enc : \mathcal{D} \rightarrow \mathcal{E}$ be our protected victim encoder that maps data points from the input to the embedding space.
Assume user $U$ has, so far, queried a subset of their data points ${q_1,\dots,q_j}$ with $j\leq n$ to the encoder and obtained the representations ${r_1,\dots,r_j}$ with each $r_i \in \mathbb{R}^s$.
We aim to estimate the true fraction of the embedding space $\efra^U$ that is covered by all returned representations ${r_1,\dots,r_j}$ to user $U$ and denote our estimate by $\tilde{\mathcal{E}}_f^U$.

\paragraph{Local Sensitive Hashing.}
One of the methods to approximate the occupied space by representations returned to a given user is via Local Sensitive Hashing (LSH)~\cite{slaney2008locality}. 
We rely on this approach for the concrete instantiation of our B4B and use it to track per-user coverage of the embedding space.
%\tf{If we think we'll be able to give an overview over other possibilities and have small results, we can say something like "We present alternatives in the Appendix."}
Standard (cryptographic) hash functions are characterized by high dispersion such that hash collisions are minimized. 
In contrast, LSH hashes similar data points into the same or proximal, so-called \textit{hash buckets}. 
This functionality is desired when dealing with searches in high-dimensional spaces or with a large number of data points. 
Formally, an LSH function $\mathcal{H}$ is defined for a metric space $\mathcal{M}=(M, d)$, where $d$ is a distance metric in space $M$, with a given threshold $T > 0$, approximation factors $f>1$, and probabilities $P_{1}$ and $P_{2}$, where $P_1 \gg P_2$. $\mathcal{H}$ maps elements of the metric space to buckets $b\in B$ and satisfies the following conditions for any two points $q_1,q_2 \in M$: (1) If $d ( q_1 , q_2 ) \le T$, then $\mathcal{H}(q_1) = \mathcal{H} ( q_2 )$ (\ie $q_1$ and $q_2$ collide in the same bucket $b$) with probability at least $P_{1}$. (2) If $d ( q_1 , q_2 ) \ge fT$, then $\mathcal{H}(q_1) = \mathcal{H}( q_2 )$ with probability at most $P_{2}$. 
%In \Cref{app:alternative_instantiations}, we propose alternative tracking tools. 

\subsection{Building Block 2: Cost Function Design}
\label{sub:block2cost}

Once we can estimate the coverage of an embedding space for a given user $U$ as $\tilde{\mathcal{E}}_f^U$, we need to design a cost function $\mathcal{C}: \mathbb{R^+} \rightarrow \mathbb{R^+}$ that maps from the estimated coverage to a cost.
The cost function needs to be designed such that it does not significantly penalize legitimate users while imposing a severe penalty on adversaries to effectively prevent the encoder from being stolen. 
The semantics of the cost function's range depend on the type of costs that the defender wants to enforce.
We discuss a broad range of options in \Cref{app:alternative_instantiations}.
These include monetary cost functions to adaptively charge users on a batch-query basis depending on their current coverage, costs in the form of additional computation that users need to perform in order to obtain their representations, similar to the proof of work in~\cite{powDefense}, costs in the form of delay in the interaction with the encoder behind the API~\citep{PoET2}, or costs in form of disk space that needs to be reserved by the user (similar to a proof of space~\cite{dziembowksi2013proofSpace,dziembowski2015proofs}). 
%\franzi{@Adam, do you have a citation here? And others you want to add?}
Which type of cost function is most adequate depends on the defender's objective and setup.
%, and \franzi{@Adam, another word for the enumeration here?}.

\paragraph{Exponential Cost Functions to Adjust Utility of Representations.}
In the concrete instantiation of B4B that we present in this work, we rely on costs in the form of the utility of the returned representations.
We choose this concrete instantiation because it is intuitive, effective, and can be directly experimentally assessed.
Moreover, it is even suitable for public APIs where, for example, no monetary costs are applicable. 
%Additionally, the noise addition can be applied to public APIs, while the higher monetary costs cannot. pplied to public APIs (where no higher monetary cost can is possible to employed). 
%\tf{@Adam, do we want to justify our choice? If so, can you help make the statement more convincing? Adam: additionally, the noise addition can be applied to public APIs, while the higher monetary cost can't.}  
We adjust utility by adding Gaussian noise with different standard deviation $\sigma$ to the returned representations.
Since we do not want to penalize legitimate users with small coverage but make stealing for adversaries with growing coverage increasingly prohibitive, we instantiate an exponential cost function that maps from the fraction of hash buckets occupied by the user to a value for $\sigma$.
% We choose the general form of this function as 
% \begin{equation}
% \label{eq:generic_exp_func}
% f_{\lambda,\alpha}(\tilde{\mathcal{E}}_f^U)=\lambda \times (\exp^{\ln{\alpha}\times \tilde{\mathcal{E}}_f^U})-\lambda
% \end{equation}
% where $\lambda$ calibrates the strength of penalty and should be set low in order to obtain low function values for small fractions of occupied buckets and $\alpha$ is a hyper-parameter that can be tuned to make concrete fractions of occupied buckets correspond to some target $\sigma$.
% We present how to concretely tune $\lambda$ and $\alpha$ for given encoders in the experimental section (\ref{sec:eval}).
% For those reasons, we take an exponential function given by \ref{eq:generic_exp_func} as the cost function. Setting $\lambda = 0.01$ ensures that low percentages of occupied buckets translate into low function values. Then, the parameter $\alpha$ is a hyper-parameter that  corresponds to $f(b)$ = 10. The value of alpha should be equal to the desired maximal percentage of buckets that can be occupied by a legitimate user.
% and $\alpha$ to 
We choose the general form of this function as 
\begin{equation}
\label{eq:generic_exp_func}
f_{\lambda,\alpha,\beta}(\tilde{\mathcal{E}}_f^U)=\lambda \times (\exp^{\ln\frac{\alpha}{\lambda}\times \tilde{\mathcal{E}}_f^U\times\beta^{-1}}-1)
\end{equation}

where $\lambda<1$ compresses the curve of the function to obtain low function values for small fractions of occupied buckets, and then we set a target penalty $\alpha$ for our cost function at a specified fraction of filled buckets $\beta$.
For instance, if we want to enforce a $\sigma$ of $5$ at $90\%$ of filled buckets (\ie for $\tilde{\mathcal{E}}_f^U=0.9$), we would need to set $\alpha=5$ and $\beta=0.9$.

\subsection{Building Block 3: Per-User Representation Transformations against Sybil Attacks}
\label{sub:block3transformations}
Given that our defense discourages users from querying with a wide variety of data points from different distributions, an adversary could create multiple fake user accounts (sybils) and query different data subsets with more uniform representations from each of these accounts.
By combining all the obtained representations and using them to train a stolen copy, the adversary could overcome the increased costs of stealing. 
To defend against such sybil attacks, we propose individually transforming the representations on a per-user level before returning them. %This prevents the adversary from directly merging representations obtained through multiple accounts to train a local stolen copy of the encoder. 
As a result, the adversary would first have to map all the representations to one single unified space before being able to jointly leverage the representations from different accounts for their stolen copy.

Formally, for a given query $q_i$, the protected victim encoder produces a representation $r_i=f_v(q_i)$, which is transformed by a user-specific transformation $T_U(r_i)$ before being returned to the querying user $U$.
For a new user $U$, the defense randomly selects the transformation $T_U$ from all possible choices. Note that the randomness is also added on a per-transformation basis, instead of only on the level of selecting the transformations. For example, a permutation of the elements in the output representations should be different for each user. 
%Additionally, the transformations can be composed, for instance, we can combine shuffling with padding and an affine transformation. This further prevents an attacker from the possibility of simply guessing what transformation was used for a given user and reversing the applied transformations to obtain the original representations.

We formulate two concrete requirements for the transformations. First, they should preserve utility for legitimate users on their downstream tasks, and second, they should be costly to reverse for an adversary.

\paragraph{Utility Preserving Transformations.}
As a concrete instantiation for our B4B, we propose a set of transformations that fulfill the above-mentioned two requirements: (1) \textit{Affine} where we apply affine transformations to representations, (2) \textit{Pad} where we pad representations with constant values, (3) \textit{Add} where we add constant values at random positions within representations, (4) \textit{Shuffle} where we shuffle the elements in the representation vectors, and (5) \textit{Binary} where the original representations are mapped to binary vectors relying on a random partitioning of the representation space. 
To preserve the full amount of information contained in the original representations, in our binary transformations, we tune the length of binary representations. 
We visualize the operation of each of these transformations in \Cref{app:alternative_instantiations}.
All these transformations can additionally be combined with each other, which further increases the possible set of transformations applied per user.
This renders it impossible for an adversary to correctly guess and reverse the applied representation.
%While standard affine transformations or shuffling of the elements fulfill the first requirement, we observe that they can be efficiently reversed.
%Therefore, we develop a transformation scheme that relies on partitioning the representation space.
Instead, the adversary has to remap the representations over all accounts into a single embedding space in order to unify them and leverage them for training of their stolen encoder copy.
%Sybil adversaries could try to circumvent the defense in different ways to unify the transformed representations into a single space (
We present an exhaustive list of strategies that adversaries can apply for the remapping in \Cref{app:sybils}. 
%The Sybil adversaries might employ many different strategies, which we present fully in the appendix. 
%Sybils use many fake user accounts to deploy their attacks, however, we show that our defense cuts such attempts short by ensuring that the remapping between representations is prohibitive even for a pair of users.
All the attack methods reduce to the minimum of remapping between representations of a pair of users, \ie they are at least as complex as mapping between two separate accounts. 
In the next section, we show that our defense already impedes stealing for an adversary with two accounts.

\section{Empirical Evaluation}
\label{sec:eval}
We first empirically evaluate our instantiation of B4B's three building blocks and show how to calibrate each of them for our defense. %\tf{Needs to be adjusted according to what we really do!}
Finally, we provide an end-to-end evaluation that highlights B4B's effectiveness in preserving downstream utility for legitimate users while successfully preventing the stealing by adversaries.

\paragraph{Experimental Setup.}

% - MODEL X, MODEL Y from reference, trained with ImageNet.
% - dataset that we consider as downstream data: CIFAR10, SVHN, STL10, and F-MNSIT

We conduct experiments on various kinds of downstream tasks and two popular SSL encoders.
To test our defense, we use 
%MNIST, \franzi{I took it out because it is confusing, since never mentioned in the main paper.}
FashionMNIST, SVHN, STL10, and CIFAR10 as our downstream datasets, each with standard train and test splits.
For stealing, we utilize training data from ImageNet and LAION-5B.
We rely on encoder models from the SimSiam~\citep{SimSiam} and the DINO~\citep{dino_2021_ICCV} SSL frameworks.
As our victim encoders, we use the publicly available ResNet50 model from SimSiam trained for 100 epochs on ImageNet and the ViT Small DINO encoder trained for 800 epochs on ImageNet, each using batch size 256. The ViT architecture takes as input a grid of non-overlapping contiguous image patches of resolution $N$x$N$. In this paper, we typically use $N = 16$.
%The output dimension of SimSiam is 2048 
The Simsiam encoder has an output representation dimension of 2048, while DINO returns 1536 dimensional representations.
%We use representations of dimension equal to 2048 for the former, and 1536 for the latter model.
We examine the utility of downstream classifiers using SimSiam's or DINO's representations obtained for the respective downstream datasets. 
To implement LSH, we rely on random projections~\cite{pauleve2010locality} that we implement from scratch. %using \franzi{@Janek: PYTHON LIBRARY}.
For a detailed description of our stealing and downstream training parameters, we refer to \Cref{app:epxeriments}.
%More details on the experimental setup are in \Cref{}.

%\franzi{@Adam, give it a look and put all things that you still think are important.}

%\name uses \adam{add the name of the library used},\janek{LSH via random projection\cite{pauleve2010locality}} which fulfills the above criteria and the number of occupied buckets provides a good approximation of the representation space coverage. \janek{In this technique, each vector is multiplied by a random projection matrix to obtain a lower-dimensional representation. Then, each component of the resulting projection is thresholded to obtain a binary hash. To ensure stable results we average the number of occupied buckets for 5 random projections.}

\begin{figure}[t]
\vspace{-0.5cm}
        \begin{subfigure}[b]{0.5\textwidth}
            \centering
            \includegraphics[width=1.0\textwidth, trim={0cm 0cm 0cm 0cm}]{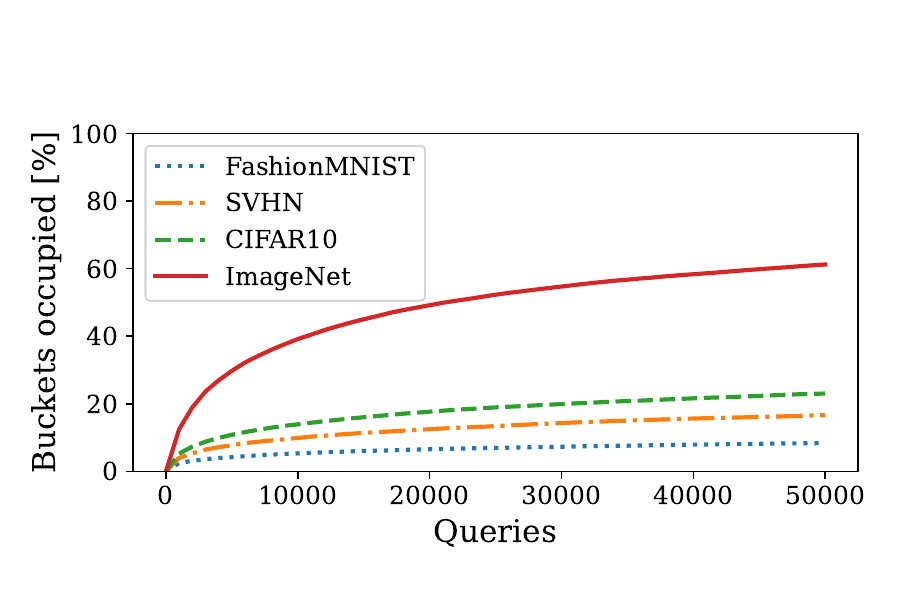}
            \vspace{-0.9cm}
            \caption[]%
            {{\small Fraction of Occupied Buckets.}}    
            \label{fig:bucket-query-number}
        \end{subfigure}
        \begin{subfigure}[b]{0.5\textwidth}
            \centering
            \includegraphics[width=1.0\textwidth, trim={0cm 0cm 0cm 0cm}]{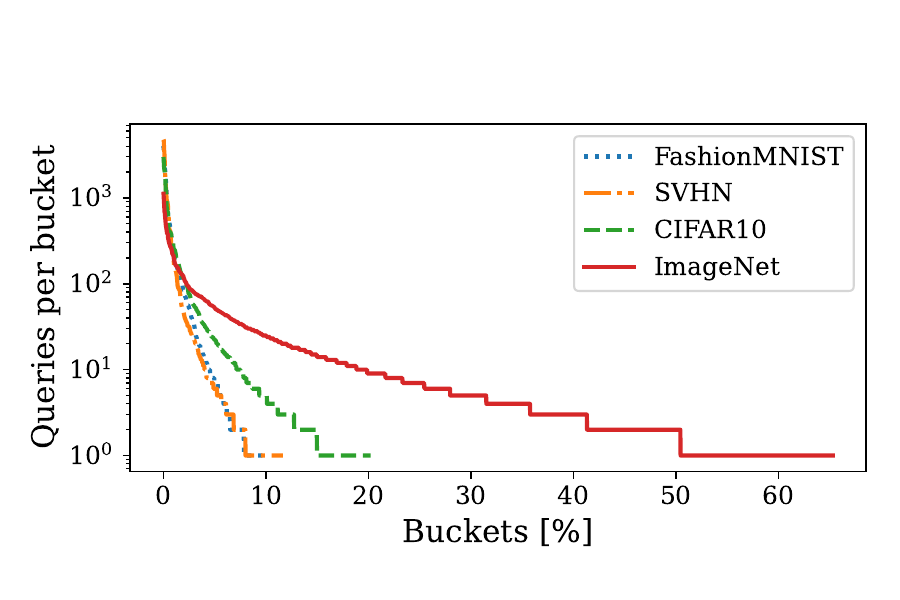}
            \vspace{-0.9cm}
            \caption[]%
            {{\small Number of Queries per Bucket.}}    
            \label{fig:bucket-histogram}
        \end{subfigure}
    \vspace{-0.1in}
    \caption{\textbf{Estimating Embedding Space Coverage through LSH on SimSiam Encoder.} 
    We present the fraction of buckets occupied by representations of different datasets as a function of the number of queries posed to the encoder \textit{(left)}.
    We observe that representations for the downstream datasets (FashionMNIST, SVHN, CIFAR10) occupy a smaller fraction of buckets than representations from the complex ImageNet dataset.
    Our evaluation of the number of queries whose representations are mapped to the same bucket \textit{(right)} indicates that our total number of buckets ($2^{12}$) is well calibrated for the estimation of covered representation space: over all datasets, we experience hash collisions, \ie queries whose representations are mapped to the same buckets. This indicates that our LSH is capable of representing similarities in the representations.
    %\franzi{What Model? SimSiam -- i"ll integrate this text.}
    %We can identify malicious vs legitimate queries by measuring the disparity of their representations in the embedding space of an encoder.
    %The queries from the same downstream task distribution are clustered together in fewer buckets than queries from different downstream tasks. 
    %(a) With more queries, more buckets are used, but the number of buckets occupied by malicious queries grows much faster than for legitimate users.
    %(b) The histogram of buckets for different users. The x-axis represents the bucket id and the y-axis is the frequency of how many representations fall into a given bucket. \janek{This is no longer on the plot:} For legitimate users (1 and 2, shown as black and blue lines, respectively), the occupied buckets are clustered together, whereas the attacker's histograms are flat since their representations fall evenly across many buckets.
    %We show that the representations for a given task are adjacent whereas representations from distinct tasks are far apart from each other.
    }
    \label{fig:bucket-query-number}
\end{figure}

\subsection{Local Sensitive Hashing for Coverage Estimation}

We first observe that the choice of the total number of hash buckets in the LSH influences the effectiveness of our method.
In the extreme, if we have a too large number of buckets, the number of buckets filled will correspond to the number of queries posed by a user which fails to capture that similar representations cover similar sub-spaces of the embedding space, and hence does not serve to approximate the total fraction of the embedding space covered.
However, if we have too few buckets, even the representations for simple downstream tasks will fill large fractions of buckets, making it impossible to calibrate the cost function such that it only penalizes adversaries.
We experimentally find that for our evaluated encoders, $2^{12}$ buckets represent a good trade-off.
In \Cref{app:set-number-of-buckets}, we present an ablation study on the effect of the number of total buckets.

Our evaluation of the LSH to track coverage of the embedding space is presented in \Cref{fig:bucket-query-number}.
We observe that representations returned for standard downstream tasks (FashionMNIST, SVHN, CIFAR10) occupy a significantly smaller fraction of the total number of buckets than complex data from multiple distributions (ImageNet, LAION-5B). 
We present additional experimental results on measuring the coverage of the representation space in \Cref{app:coverage-space}.
Specifically, we show that our method of measuring the embedding space coverage has broad applicability across various encoders and datasets used for pretraining.
% Specifically, when querying the encoder with ImageNet-Full (includes all 1000 classes) and LAION-5B datasets, they both occupy a large fraction of the representation space of the victim encoder, as shown in Figure \ref{fig:imagenet}. In contrast, CIFAR10 covers the smallest portion of the representation space as the simplest dataset tested. ImageNet-Dogs (with only 118 classes for dog breeds) falls in the middle, occupying more space than CIFAR10 but less than ImageNet-Full and LAION-5B. Its intermediate coverage aligns with its mid-level difficulty compared to the other datasets. As indicated by representation space coverage, stealing the encoder is similarly effective with ImageNet-Full and LAION-5B datasets, as both datasets cover a large fraction of the representation space. Overall, Figure \ref{fig:imagenet} demonstrates that: 1) our \name can successfully protect the encoder model even from attackers stealing with data that was not used to train the model (LAION-5B in this case) and 2) while providing clean representation for users querying from downstream tasks that are part of more complicated datasets (ImageNet-Dogs).
We further observe that the fraction of buckets occupied by the representations saturates over time. %e, bounding the penalty that will be added to legitimate users.
These findings highlight that LSH is successful in capturing the differences between legitimate users and adversaries---even in a low-query regime.
Finally, we note that our total number of buckets ($2^{12}$) is well calibrated since, over all datasets, it successfully maps multiple representations to the same hash bucket while still filling various fractions of the total number of buckets.

\subsection{Calibrating the Cost Function}
\begin{wrapfigure}{r}{6cm}
    \vspace{-1cm}
    \centering
        \includegraphics[width=0.4\textwidth, trim={0cm 0.5cm 0cm 0cm},clip]{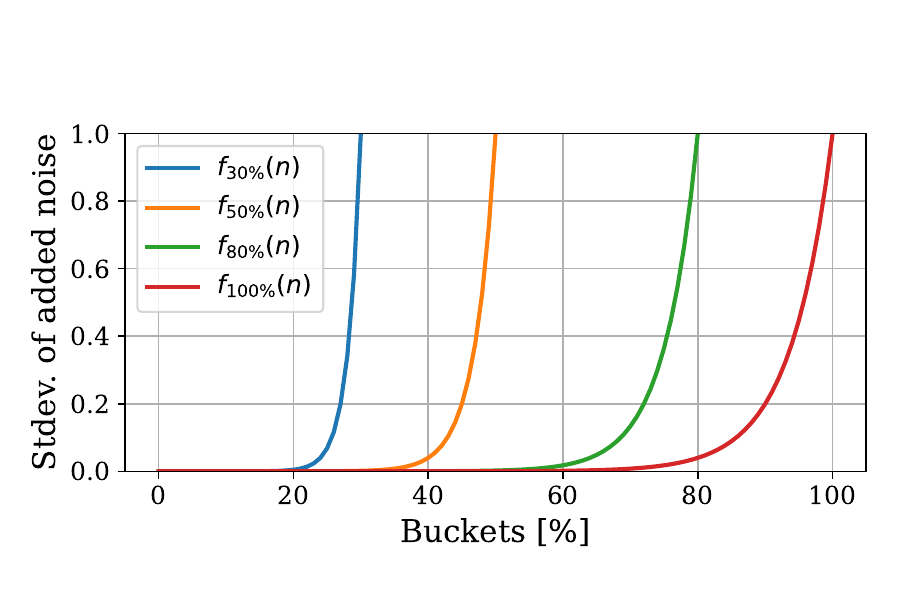}
    %\vspace{-2ex}
    %\caption{\textbf{Lower Dimension Representation of Different Representations.} We map the representations obtained for different downstream tasks to a two-dimensional space. We observe that the different downstream tasks form clusters.}
    \caption{\textbf{Cost Function Calibration.} 
    %The higher the disparity of query representations, the higher the cost of querying the encoder. 
    %Legitimate users operate in the regime of low query cost, while users who query from many data distributions incur much higher costs.
    %We show that the representations for a given task are adjacent whereas representations from distinct tasks are far apart from each other.
    }
\vspace{-0.5cm}
    \label{fig:cost-bucket}
\end{wrapfigure}
We experiment with different sets of hyperparameters to instantiate the cost function from \Cref{eq:generic_exp_func} in the previous section (\ref{sub:block2cost}). %$f_{\lambda,\alpha}(\tilde{\mathcal{E}}_f^U)=\lambda \times (\exp^{\ln{\alpha}\times \tilde{\mathcal{E}}_f^U})-\lambda$ that we introduced in the prior section. 
%\tf{Franzi: Probably, we would also need some form of small ablation in the appendix.}
As described there, we can calibrate the function (as shown in \Cref{fig:cost-bucket}) such that a desired penalty (in the form of a specific $\sigma$) will be assigned at a certain fraction of buckets occupied.
For B4B, we aim at penalizing high embedding space coverage severely.
Therefore, we need to identify and optimize for two components: 1) which value of $\sigma$ leads to significant performance drops, and 2) for what fraction of coverage do we want to impose this significant drop.
We base both components on empirical observations. Our first observation is that for our four downstream tasks (FashionMNIST, SVHN, STL10, and CIFAR10), performance drops to 10\% (\ie random guessing) at roughly $\sigma=0.5$.
In \Cref{fig:bucket-query-number}, we further see that with 50k queries, the downstream tasks occupy $<30\%$ of the buckets.
Ultimately, setting $\alpha$ and $\beta$ are design choices that an API provider needs to make in order to specify what type of query behavior they want to penalize.
As very loose bounds (weak defense), based on our observation, we consider $\sigma=1$ as a high penalty, which leads to $\alpha=1$, and select $\beta=0.8$. This $\beta$ corresponds to considering 80\% of buckets filled as a too-large coverage of the embedding space. We empirically observe that coverage of 80\% of buckets occurs, for example, after around 100k of ImageNet queries.
By choosing our target $\beta$ so loose, \ie significantly larger than the observed $30\%$ for downstream tasks, we offer flexibility for the API to also provide good representations for more complex downstream tasks.
% and needs to be calibrated according to the given encoder and desired strength of the defense.
% As a loose bound, we therefore, choose $\sigma=1$ as costs that we consider destructive
% \franzi{Come up with something here. The only additional justifaction for such a low penalization is that if we overfit the cost function to our downstream task there is no gap for adding a new legit task}
% As a result, in the final cost function, we aim for a $\sigma=1$ which specifies our parameter $\alpha$ to be $1$.
% The second question is a design choice by the API provider that specifies what type of query behavior they want to penalize and needs to be calibrated according to the given encoder and desired strength of the defense.
% In \Cref{fig:cost-bucket} we depict multiple instantiations of our cost function that differ only in the targeted fraction of coverage $\beta$ where the severe penalty of $\sigma$ should apply.
% Since we experiment with a SimSiam encoder trained on ImageNet, we would like to penalize querying with similarly complex datasets.
% From \Cref{fig:bucket-query-number}, we observe that querying with ImageNet occupies increasing number of buckets. 
% As a rather loose cost function, we select $\beta=0.8$ as a target. This corresponds to 80\% of buckets filled, which occurs after around 100k ImageNet queries.
Finally, to obtain a flat cost curve close to the origin---which serves to map small fractions of covered buckets to small costs---we find that we can set $\lambda=10^{-6}$.
In the Appendix, we evaluate our defense end-to-end with differently parameterized cost functions.

\subsection{Assessing the Effect of Transformations}

\paragraph{Transformations Do Not Harm Utility for Legitimate Users.}
%We first show that our transformations do not harm utility for legitimate users.
We evaluate the downstream accuracy for transformed representations based on training a linear classifier on top of them. % on the transformed representations returned for different downstream tasks.
To separate the effect of the noise added by our defense from the effect of the transformations, we perform the experiments in this subsection without adding noise to the returned representations.  
For example, on the CIFAR10 dataset and a SimSiam encoder pre-trained on ImageNet, without any transformations applied, we obtain a downstream accuracy of 90.41\% ($\pm$ 0.02), while, with transformations, we obtain 90.24\% ($\pm$ 0.11)  for Affine, 90.40\% ($\pm$ 0.05) for Pad+Shuffle, 90.18\% ($\pm$ 0.06) for Affine+Pad+Shuffle, and 88.78\% ($\pm$ 0.2) for Binary.
%For the DINO encoder 
%\ta{We can also add numbers here for DINO.}
This highlights that the transformations preserve utility for legitimate users.
%\ta{(\eg with less than 1\% drop for Affine transformations in the case of SimSiam).}
%\tf{Add results with decimal point @Stachu. Adam: added.}
This holds over all datasets we evaluate as we show in \Cref{app:epxeriments}. 

\paragraph{Adversaries Cannot Perfectly Remap Representations over Multiple Sybil Accounts.}
To understand the impact of our per-user account transformations on sybil-attack based encoder stealing, we evaluate the difficulty of remapping representations between different sybil accounts.
For simplicity, and since we established in \Cref{sub:block3transformations} that multi-account attacks reduce to a two-account setup, we assume an adversary who queries from two sybil accounts and aims at learning to map the transformed representations from account \#2 to the representation space of account \#1.
%Note that when querying from more than two accounts the adversary eventually still has to map all representations to one single representation space.
%As a consequence, the multi-account setup, reduces to the two-account setup, 
Using more accounts for the adversary causes a larger query overhead and potentially more performance loss from remapping.
Our evaluation here, hence, represents a lower bound on the overhead caused to the adversary through our transformations.

We learn the mapping between different accounts' representations by training a linear model on overlapping representations between the accounts.
We assess the fidelity of remapped representations as a function of the number of overlapping queries between the accounts.
As a fidelity metric for our remapping, we compare the cosine distance between representations ($a$ and $b$ defined as: $1 - \frac{a^{T} b}{||a||_2 \cdot ||b||_2}$).
Once the remapping is trained, we evaluate by querying 10k data points from the test dataset through account \#1 and then again through account \#2. 
Then, we apply the learned remapping to the latter one and compute the pairwise cosine distances between the representations from account \#1 and their remapped counterparts from account \#2.
%\stachu{This setup for the extended experiment should be clarified between us and then updated here.}
\begin{wrapfigure}{r}{5.5cm}
    \vspace{-0.3cm}
    \centering
    \includegraphics[width=0.35\textwidth, trim={0cm 0.0cm 0cm 0cm},clip]{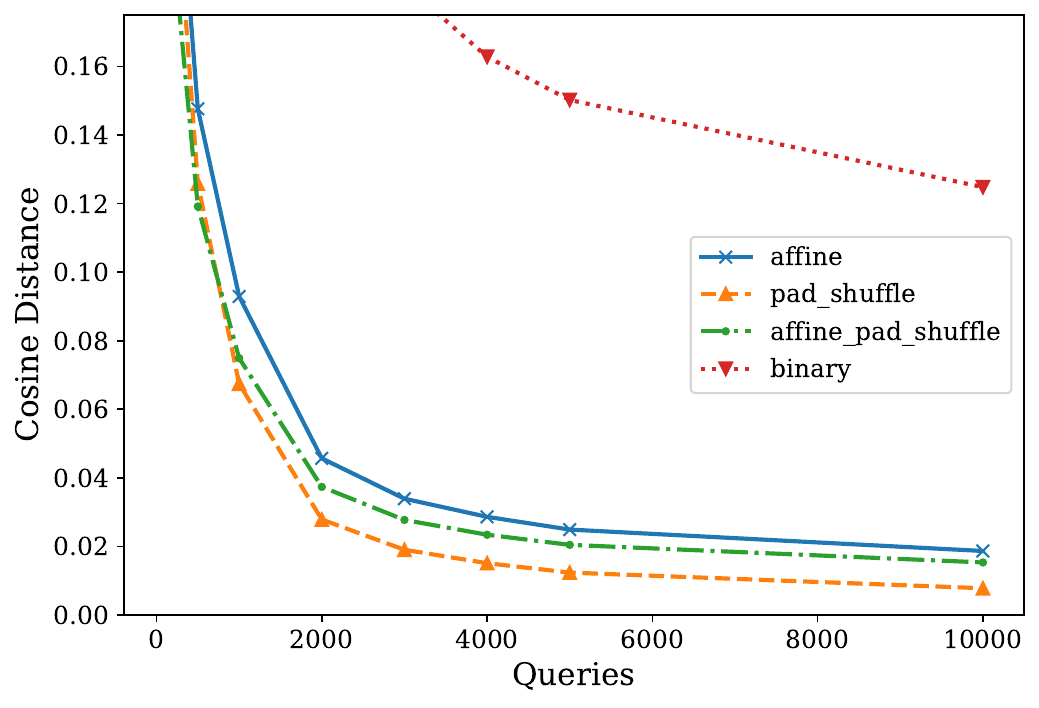}
    %\vspace{-2ex}
    %\caption{\textbf{Lower Dimension Representation of Different Representations.} We map the representations obtained for different downstream tasks to a two-dimensional space. We observe that the different downstream tasks form clusters.}
    \caption{\textbf{Quality of Remappings.} 
    %The higher the disparity of query representations, the higher the cost of querying the encoder. 
    %Legitimate users operate in the regime of low query cost, while users who query from many data distributions incur much higher costs.
    %We show that the representations for a given task are adjacent whereas representations from distinct tasks are far apart from each other.
    }
%\vspace{-0.35cm}
    \label{fig:quality-remap-main}
    \vspace{-0.5cm}
\end{wrapfigure}
Our results are depicted in \Cref{fig:quality-remap-main}.
%We assess the quality of the remapping using the cosine distance between representations $a$ and $b$ defined as: $1 - \frac{a^{T} b}{||a||_2 \cdot ||b||_2}$.
%We show that over the three datasets (MNIST, CIFAR10, and ImageNet) in the defended and undefended setup, 
We show that the largest cosine distance is achieved with the binary transformations, making them the most protective against the adversary since they best prevent perfect remapping, even with an overlap of as many as 10k queries between both accounts. However, these binary transformations also incur the highest drop in accuracy for legitimate users. The defender has the possibility of selecting their preferred types of transformations between representations taking into account the trade-offs between the effectiveness of the defense and the negative impact on legitimate users. 

\subsection{End-to-End Stealing of an Encoder under our Defense}

We perform an end-to-end study to showcase how our \name defense affects legitimate users vs adversaries. 
The hyperparameters for B4B are chosen according to the empirical evaluation of the previous sections with $2^{12}$ as the number of buckets, $\alpha=1, \beta=0.8, \lambda=10^{-6}$ as the hyperparameter of the cost function, and different random affine transformations per-user account.
Our main results are presented in \Cref{tab:StealImagenet}.  
We observe that instantiating our framework with \name has a negligible impact on legitimate users while substantially lowering the performance of stolen encoders in the case of single-user and sybil attackers.

\definecolor{mygray}{gray}{0.5}
\addtolength{\tabcolsep}{-3pt} 
\begin{table}[t]
\caption{\textbf{Stealing and Using Encoders With and Without our Defense.} 
%We depict the performance of the victim encoder for legitimate users, and the performance of the stolen encoders on various downstream tasks.
%Then row presents the performance of the stolen encoder with 50k queries without our defense.
%In the rows under the dashed line, we present the performance of the encoder stolen from the victim with our defense.
%We observe that 
%\franzi{@Janek to fill}
%\janek{the results for SVHN look strange - the downstream model gets stuck on acc@1 19.6 after 1 epoch of training and acc@5 only slightly oscillates. This is the same for both stolen encoders with defense. I tested the accuracy for the reference encoder stolen with 50k and no defense and for that model I got normal acc.}
The \textit{USER} column represents the type of the APIs' user, where {\footnotesize LEGIT} denotes a legitimate user, {\footnotesize ATTACKER} stands for a standard single-account adversary, and {\footnotesize SYBIL} represents an adversary using two sybil accounts. We use InfoNCE loss for encoder extraction. \# Queries stands for the number of queries used for stealing with {\footnotesize ALL} denoting that the entire downstream dataset was used.
The \textit{TYPE} column expresses how the dataset is used.
We follow the stealing setup from~\citep{SSLdatasetinference}. 
In the first row, we present the undefended victim encoder's performance as the accuracy for downstream tasks trained on the encoder's returned representations. 
In the following row, we show downstream utility for legitimate users when the victim encoder is defended by our \name.
%Then, we present accuracy on downstream tasks for legitimate users, and finally, 
Finally, (in the remaining rows) we assess the performance of stolen encoders on the downstream tasks.
Our results highlight that while the performance of the encoder for legitimate users stays high, our \name renders stealing inefficient with the stolen encoders obtaining significantly worse performance on downstream tasks.
%\janek{The values in the top row are for datasets limited to 50k (because the following rows are also limited to 50k. This affects only SVHN 74.98 50k and 79.01 full dataset}adam{I changed the number of queries.}
}
\label{tab:StealImagenet}
\begin{center}
\footnotesize %franzi: we are not allowed to go over borders
\begin{sc}
\begin{tabu}{ccccccccc}
\toprule
User&Defense& \# Queries & Dataset & Type & CIFAR10 &  STL10 & SVHN & F-MNIST\\
\midrule
% N/A&N/A& \textit{Victim} & \textit{ImageNet} & \textit{Train}  & 90.41{\tiny $\pm$0.02} & 95.08{\tiny$\pm$0.13} & 75.47{\tiny $\pm$0.04} & 91.22{\tiny $\pm$0.11} \\
% % N/A&N/A& \textit{Victim} & N/A & \textit{ImageNet} & \textit{Train} & 90.49 &  94.9 & 74.98 & 90.57 \\
% \cdashlinelr{1-9}

\rowfont{\color{mygray}} \Legit & None &  All &  Task & Query & 90.41 {\tiny $\pm$0.02} & 95.08{\tiny $\pm$0.13} & 75.47{\tiny $\pm$0.04} & 91.22{\tiny $\pm$0.11}\\ 	
%\rowfont{\color{mygray}} \Legit & None &50k & N/A  & STL10  & N/A &  & N/A & N/A\\
 % \Legit &  $\sigma$=0.1 &  All &   Task &  Query & 90.20{\tiny $\pm$0.03}& 95.15{\tiny $\pm$0.13} 	 &75.29{\tiny $\pm$0.09}& 91.24{\tiny $\pm$0.02}\\	 	
%\rowfont{\color{mygray}} \Legit & None &50k & N/A  & SVHN  & N/A & N/A &  & N/A\\
 % \Legit &  $\sigma$=10 &  All&   Task &  Query &65.11{\tiny $\pm$0.45} & 76.37{\tiny $\pm$0.14} &33.23{\tiny $\pm$0.09}& 65.83{\tiny $\pm$0.13} \\ 	 	 	
%\rowfont{\color{mygray}} \Legit & None &50k & N/A  &  F-MNIST   & N/A & N/A & N/A &\\
 \Legit &  \name &  All &   Task  &  Query  & 90.24{\tiny $\pm$0.11}  & 95.05{\tiny $\pm$0.1} & 74.96{\tiny $\pm$0.13} & 91.7{\tiny $\pm$0.01}\\	 	 	
%\name &50k (single user)& N/A  & STL10   \\
%\name &60k (single user)& N/A  & F-MNIST   \\
\cdashlinelr{1-9}
\rowfont{\color{mygray}}  \Attacker &  None & 50K & ImgNet &  Steal & 65.2{\tiny $\pm$0.03} & 64.9{\tiny $\pm$0.01} & 63.1{\tiny $\pm$0.01}  & 88.5 {\tiny $\pm$0.01}\\ 
 % \Attacker &  $\sigma$=0.1 &50k &ImgNet  &  Steal  & 64.92{\tiny $\pm$0.04} & 64.61{\tiny $\pm$0.02} & 62.35{\tiny $\pm$0.01}  & 88.41{\tiny $\pm$0.01} \\ 	 	 	
 % \Attacker &   $\sigma$=10 & 50K & ImgNet &  Steal  &36.32{\tiny $\pm$0.2}  &32.59{\tiny $\pm$0.06} & 20.59{\tiny $\pm$0.01}& 74.94{\tiny $\pm$0.02} \\	 	 	
%\cdashlinelr{1-9}
 \Attacker & \name &50k & ImgNet & Steal  & 35.72{\tiny $\pm$0.04}  & 31.54{\tiny $\pm$0.02} & 19.74{\tiny $\pm$0.02}  & 70.01{\tiny $\pm$0.01} \\

% %\hdashline
% %\rowfont{\color{mygray}} \Legit &None &50k & N/A  & CIFAR10  &  & N/A & N/A & N/A\\
% \Legit &\name &50k & CIFAR10 & Query & 90.24{\tiny $\pm$0.11} & N/A & N/A & N/A\\
% %\rowfont{\color{mygray}} \Legit & None &50k & N/A  & STL10  & N/A &  & N/A & N/A\\
% \Legit & \name &5k & STL10 & Query & N/A & 95.05{\tiny $\pm$0.1} & N/A & N/A\\
% %\rowfont{\color{mygray}} \Legit & None &50k & N/A  & SVHN  & N/A & N/A &  & N/A\\
% \Legit & \name &73k & SVHN & Query & N/A & N/A & 74.96{\tiny $\pm$0.13} & N/A\\
% %\rowfont{\color{mygray}} \Legit & None &50k & N/A  &  F-MNIST   & N/A & N/A & N/A &\\
% \Legit & \name &60k &  F-MNIST & Query  & N/A & N/A & N/A & 91.7{\tiny$\pm$0.01}\\
% %\name &50k (single user)& N/A  & STL10   \\
% %\name &60k (single user)& N/A  & F-MNIST   \\
% \cdashlinelr{1-9}

% \rowfont{\color{mygray}} \Attacker & None & 50K & ImageNet & Steal & 65.2 {\tiny $\pm$0.03} & 64.9 {\tiny $\pm$0.01} & 62.1 {\tiny $\pm$0.01} & 88.5 {\tiny $\pm$0.01}\\
% \Attacker &\name &50k & ImageNet & Steal  & 35.72{\tiny$\pm$0.04} & 31.54{\tiny $\pm$0.02} & 19.74\tiny{$\pm$0.02} & 70.01{\tiny$\pm$0.01} \\

\rowfont{\color{mygray}} \Attacker & None & 100K & ImgNet & Steal  & 68.1 {\tiny $\pm$0.03} & 63.1 {\tiny $\pm$0.01} & 61.5 {\tiny $\pm$0.01} & 89.0 {\tiny $\pm$0.07}\\
%\cdashlinelr{1-9}
\Attacker &\name &100k & ImgNet & Steal  & 12.01{\tiny$\pm$0.07} & 13.94{\tiny$\pm$0.05} & 19.96{\tiny$\pm$0.03} & 69.63{\tiny$\pm$0.07} \\
\rowfont{\color{mygray}} \Attacker & None & 100K &  LAION & Steal  & 
64.92{\tiny $\pm$0.03} 	& 62.51{\tiny $\pm$0.03} 	& 59.02{\tiny $\pm$0.02} 	& 84.54{\tiny $\pm$0.01} \\
% 68.1 {\tiny $\pm$0.03} & 63.1 {\tiny $\pm$0.01} & 61.5 {\tiny $\pm$0.01} & 89.0 {\tiny $\pm$0.07}\\
%\cdashlinelr{1-9}
\Attacker &\name &100k & LAION & Steal & 
40.96{\tiny$\pm$ 0.03} 	& 40.69{\tiny$\pm$ 0.05} & 	34.43{\tiny$\pm$ 0.01} 	& 72.92{\tiny$\pm$ 0.01} \\
%12.01{\tiny$\pm$0.07} & 13.94{\tiny$\pm$0.05} & 19.96{\tiny$\pm$0.03} & 69.63{\tiny$\pm$0.07} \\
% \cdashlinelr{1-8}
% Our Defense (sybil) &25k + 25k& InfoNCE  & ImageNet   & & & & \\
\cdashlinelr{1-9}
\Sybil &\name &2$\times$50k & ImgNet & Steal  & 39.56{\tiny$\pm$ 0.06} & 38.50{\tiny$\pm$0.04} & 23.41{\tiny$\pm$0.02} & 77.01{\tiny$\pm$ 0.08} \\
\Sybil &\name &3$\times$33.3k & ImgNet & Steal  & 33.87{\tiny$\pm$ 0.05} & 38.57{\tiny$\pm$0.06} & 21.16{\tiny$\pm$0.01} & 72.95{\tiny$\pm$ 0.05} \\
\Sybil &\name &4$\times$25k & ImgNet & Steal  & 33.98{\tiny$\pm$ 0.04} & 34.52{\tiny$\pm$0.08} & 21.21{\tiny$\pm$0.02} & 70.71{\tiny$\pm$ 0.05} \\
\Sybil &\name &5$\times$20k & ImgNet & Steal  & 32.65{\tiny$\pm$ 0.05} & 32.45{\tiny$\pm$0.05} & 29.63{\tiny$\pm$0.01} & 70.12{\tiny$\pm$ 0.08} \\
\Sybil &\name &6$\times$16.7k & ImgNet & Steal  & 26.62{\tiny$\pm$ 0.04} & 26.85{\tiny$\pm$0.05} & 24.32{\tiny$\pm$0.02} & 70.51{\tiny$\pm$ 0.04} \\
\bottomrule
\end{tabu}
\end{sc}
\end{center}
\end{table}
\addtolength{\tabcolsep}{3pt} 

\textbf{Legitimate Users.} We compare the accuracy of downstream classifiers trained on top of unprotected vs defended encoders. The victim encoder achieves high accuracy on the downstream tasks when no defense is employed. 
%Then, we apply \name and measure the same accuracy levels. 
With \name in place, we observe that across all the downstream tasks, the drop in performance is below 1\%. For example, there is only a slight decrease in the accuracy of CIFAR10 from 90.41$\pm0.02$\% to 90.24$\pm0.11$\%.
\name's small effect on legitimate users stems from the fact that their downstream representations cover a relatively small part of the representations space.
This results in a very low amount of noise added to their representations which preserves performance. 

\textbf{Adversaries.} 
For adversaries who create a stolen copy of the victim encoder, we make two main observations.
The most crucial one is that when our \name is in place, the performance of the stolen copies over all downstream tasks significantly drops in comparison to when the victim encoder is unprotected (grey rows in \Cref{tab:StealImagenet}).
This highlights that our \name effectively prevents stealing.
Our next key observation concerns the number of stealing queries used by the adversary:
When no defense is applied, the more queries are issued against the API (\eg 100k instead of 50k), the higher performance of the stolen encoder on downstream tasks (\eg CIFAR10 or FashionMNIST). 
In contrast, with \name implemented as a defense, the performance decreases when using more stealing queries from a single account.
This is because with more queries issued, the coverage of embedding space grows which renders the returned representations increasingly noisy and harms stealing performance.
%In the case of a single-user adversary, the performance of the encoder increases, however, when \name is used, the noise added to the representations significantly drops the performance of the stolen encoder. Note that we observe lower performance for 100K queries than 50K queries when \name is used since with more queries, our defense increases the amount of added noise, which distorts the whole encoder more and further lowers its performance to almost random accuracy on the downstream tasks.

Moreover, we show that \name can also prevent model stealing attacks with data from a different distribution than the victim encoder's training set. We highlight this in \Cref{tab:StealImagenet} where we also use the LAION-5B dataset to steal an ImageNet pre-trained encoder. Our results highlight first that without any defense in place, the LAION dataset is highly effective to extract the ImageNet pre-trained encoder. Second, \name effectively defends against such attacks, and yields a significant drop in downstream accuracy (on average above 20\%) of the stolen encoder. 

We also show that this performance drop cannot be prevented by sybil attacks. 
Therefore, we first consider an adversary who queries from two sybil accounts with 50k queries issued per account and the first 10k queries of both accounts used to learn the remapping of representations between them.
When the adversary trains their stolen encoder copy on all the remapped representations, they increase downstream performance over querying from a single account.
Yet, their performance is still significantly smaller than the performance of the victim encoder for legitimate users, or the encoder stolen from an undefended victim. Moreover, using more than two sybil accounts further reduces the attack performance as remapping complications accumulate. With ten sybils, remapping leaves no more usable data for training the stolen encoder. This demonstrates our method's advantage: increasing the number of sybil accounts makes encoder extraction impractical due to the growing remapping overhead. 
Overall, the results highlight that our \name also successfully prevents sybil attacks. %\tf{I as a reviewer would want to see more examples evaluated for the sybil case: more accounts, more or less queries used for the remapping. We should get as much ready for the appendix as possible, the rest for the rebuttal.}

\subsection{Baseline Comparison}
\label{sub:baseline}
Finally, we compare our \name against the current state-of-the-art baseline defense, namely a static addition of noise to all the returned representations (as proposed in  \citep{SSLextraction} (Section A.4),\citep{StolenEncoder,ContSteal}). 
For direct comparability, we evaluate the defense using the our end-to-end experiment setup from the previous section.
We present our results in Table \ref{tab:StealImagenet_noise} in \Cref{app:baseline}.
Confirming the findings from \citep{SSLextraction} our results also show that defenses that rely on static noise have the great disadvantage to harm legitimate users and attackers equally.
When adding noise with a small standard deviation of $\sigma=0.1$, we observe a negligible (<1\%) performance drop for both attackers and legitimate users.
Adding noise with a large standard deviation of, for example, $\sigma=10$, we observe that both legitimate users' and attackers' performance drops between 15\% and >40\%.
In summary, these defenses can either effectively defend stealing (but harm legitimate users), or keep utility for legitimate users high (but not defend well against stealing). In contrast, our \name is able to provide high performance for legitimate users while effectively defending the encoder against stealing attacks.
\section{Conclusions}

We design \name a new and modular active defense framework against stealing SSL encoders. All the previous approaches were either reactive, acting after the attack happened to detect stolen encoders, or lowered the quality of outputs substantially also for legitimate users which rendered such mechanisms impractical. 
%We instantiate our framework as a modular active defense against encoder stealing.
We show that B4B successfully distinguishes between legitimate users and adversaries by tracking the embedding space coverage of users' obtained representations.
%keeping track on the coverage of the embedding space by users' obtained representations.
B4B then leverages this tracking to apply a cost function that penalizes users based on the current space coverage, for instance, by lowering the quality of their outputs.
Finally, \name prevents sybil attacks by implementing per-user transformations for the returned representations.
Through our experimental evaluation, we show that our defense indeed renders encoder stealing inefficient while preserving downstream utility for legitimate users.
Our B4B is therefore a valuable contribution to a safer sharing and democratization of high-utility encoders over public APIs.

% We take a different approach and consider practical cases where API encoders are exposed to extract features for a given downstream task while adversaries who try to steal such encoders query them with a diverse set of queries from different distributions. This forms a differentiating factor between legitimate and malicious users. We analyze the incoming queries and bucketize them using local sensitive hashing. The queries which fall into more buckets and have higher coverage of the representations space incur higher cost, since they are more likely intended to extract the encoder behind the API. On the other hand, legitimate users remain unaffected. To defend against adaptive attackers who create many fake accounts, called Sybils, we design a novel technique that applies different utility-preserving transformations per user. To unify the representations from many fake users, adversary is forced to re-map them, hence incurring a high cost. This approach actively increases the cost of encoder stealing, which ultimately disincentivizes the attacker. 
\acksection
This research was supported by Warsaw University of Technology within the Excellence Initiative Research University (IDUB) programme, National Science Centre, Poland grant no 2020/39/O/ST6/01478, grant no 2020/39/B/ST6/01511, grant no 2022/45/B/ST6/02817, and in part by PL-Grid Infrastructure grant nr PLG/2023/016361. The authors applied for a CC BY license to any Author Accepted Manuscript (AAM) version arising from this submission, in accordance with the grants' open access conditions.
\bibliographystyle{plainnat}
\bibliography{main}

%\newpage
%\input{contents/checklist}

%\clearpage
\appendix 
\clearpage
\section{Broader Impacts}
\label{sec:broader-impacts}

The goal of our work is to actively defend self-supervised encoders against model stealing attacks. 
Since we are directly defending encoders, any negative societal impacts of our work are minimal. 
One potentially negative impact could be the degradation of performance for legitimate users. However, as shown in our experimental results, we are able to preserve high utility for standard users.

\section{Limitations}
\label{sec:limitations}

We show how our defense method is tuned for SimSiam and DINO. There are more types of SSL encoders that can be tested with our method. The \name defense method requires tuning the parameters, such as the number of occupied buckets that is allowed without any penalty for the cost function, or the selection of the transformations. These steps are rather difficult to automate but can be replaced with more data-driven approaches. For example, instead of designing a cost function from scratch, one could create an ML model to obtain a cost for a given occupation of the representation space. We explain more details in the \Cref{app:cost-function}.

\section{Alternative Building Blocks to Instantiate B4B}
\label{app:alternative_instantiations}

While we present a reference implementation of B4B in our work that instantiates the three building blocks with (1) Local Sensitive Hashing, (2) Utility of the Representations, and (3) a set of concrete transformations, there exists a multitude of alternatives to concretely implement our B4B framework. In the following, we present these alternatives, grouped by building block. 

\subsection{Alternative Estimation of the Coverage of Embedding Space}

We also explore alternative methods to measure the distances between representations for queries sent to an API. One of them is to apply the cosine distance (where for two representations $a$ and $b$, it is defined as: $1 - \frac{a^{T} b}{||a||_2 \cdot ||b||_2}$) since it can be measured between individual data points in a pair-wise fashion. If the total pair-wise cosine distance between representations for a given user is small, then the user queries presumably come from a single downstream task distribution. Otherwise, a user might be malicious and would like to cover a large part of the representation space, then the total pair-wise cosine distance for the user's representations would be high. Note that in this case, the cosine distance can be replaced with any other distance measure, such as the Minkowski distance. We opt for the LSH in our reference implementation, since it is much less expensive to compute than cosine distance. LSH requires only $2^{12}=4096$ buckets that can be expressed as a binary table with the same number of elements, which requires in the worst case iterating over all of them to count how many are occupied. With more than $4096$ queries sent by a given user, the computation on the LSH is sublinear $<O(n)$ with respect to the number of user queries. For the cosine distance approach, the required computation grows quadratically $O(n^2)$ with the number of queries.

\subsection{Alternative Cost Functions}
\label{app:cost-function}

The cost functions can be designed from scratch manually or learned, for example, via an ML model, such as a neural network or SVM. In our initial version, the function was designed manually, where the underlying premise is that once a specified number of buckets is occupied, the cost should grow exponentially. Instead of defining such a function or providing the high-level parameters for functions that we contributed, one could learn an ML model that for a given number of buckets occupied, it should output an estimated cost, or even directly, the desired $\sigma$ (standard deviation) of the noise added to representations. This method requires a relatively large number of data points to be provided for training the model, however, lowers the burden on a defender to either decide on the specific function or adjust its parameters. Thus, it could be more user-friendly, for example, not necessitating any mathematical background, but can be precise enough to obtain the desired behavior. 

Note that instead of adding the calibrated noise (proportional to the estimated cost) to the representations, we could rather require a given user to pay a higher monetary cost for queries that cover a large fraction of the representation space, or force a user to solve a puzzle in a form of the proof-of-work~\citep{powDefense}, wait a specified amount of time via proof-of-elapsed time (PoET)~\citep{PoET2}, or prove that a specified amount of disk space was reserved~\citep{dziembowksi2013proofSpace,dziembowski2015proofs}. For example, consider the approach with PoET. A user sends queries to the API, which we cost based on their occupation of the embedding space. The user is sent a waiting time. The users’ resource (e.g., a CPU) has to be occupied for this specific waiting time without performing any work. At the end of the specified amount of time, the user sends proof of the elapsed time, which can be easily verified by the server. PoET requires access to specialized hardware, for example, secure CPU instructions that are becoming widely available in consumer and enterprise processors. If a user does not own such hardware, proof of elapsed time could be produced using a service exposed by a cloud provider (e.g., Azure VMs that feature TEE 2). Note that if a server sends the time based on the calculated cost, the adversary might learn the cost function. Instead, the exact waiting time should be split in random \textit{subwaiting} times and sent to the user one by one. Thus, a server should rather have a few rounds of exchange with the client to incur the additional cost. 

% To reduce energy consumption, our approach can be adapted to rely on PoET (Proof-of-Elapsed-Time), as implemented by Hyperledger Sawtooth, instead of PoW. The users’ resource (e.g., a CPU) would be occupied for a specific amount of time without performing any work. At the end of the waiting time, the user sends proof of the elapsed time instead of proof of work. Both proofs can be easily verified.

% PoET reduces potential energy wastage compared to PoW but requires access to specialized hardware, namely new secure CPU instructions that are becoming widely available in consumer and enterprise processors. If a user does not have such on-premise hardware, proof of elapsed time could be produced using a service exposed by a cloud provider (e.g., Azure VMs that feature TEE 2).

% PoET reduces the potential for energy waste compared with PoW. PoET requires specialized hardware with secure CPU instructions that are becoming more common. If local hardware is unavailable, a cloud service could produce the proof of elapsed time. Overall, PoET lowers energy usage versus PoW.

%in~\cite{powDefense}, costs in the form of delay in the interaction with the encoder behind the API~\citep{PoET2}, or costs in form of disk space that needs to be reserved by the user (similar to a proof of space~\cite{dziembowksi2013proofSpace,dziembowski2015proofs}). 

\subsection{Alternative to Transformations}

\begin{figure}[h!]
    \centering 
    \includegraphics[width=\textwidth, trim={0cm 0cm 0cm 0cm}]{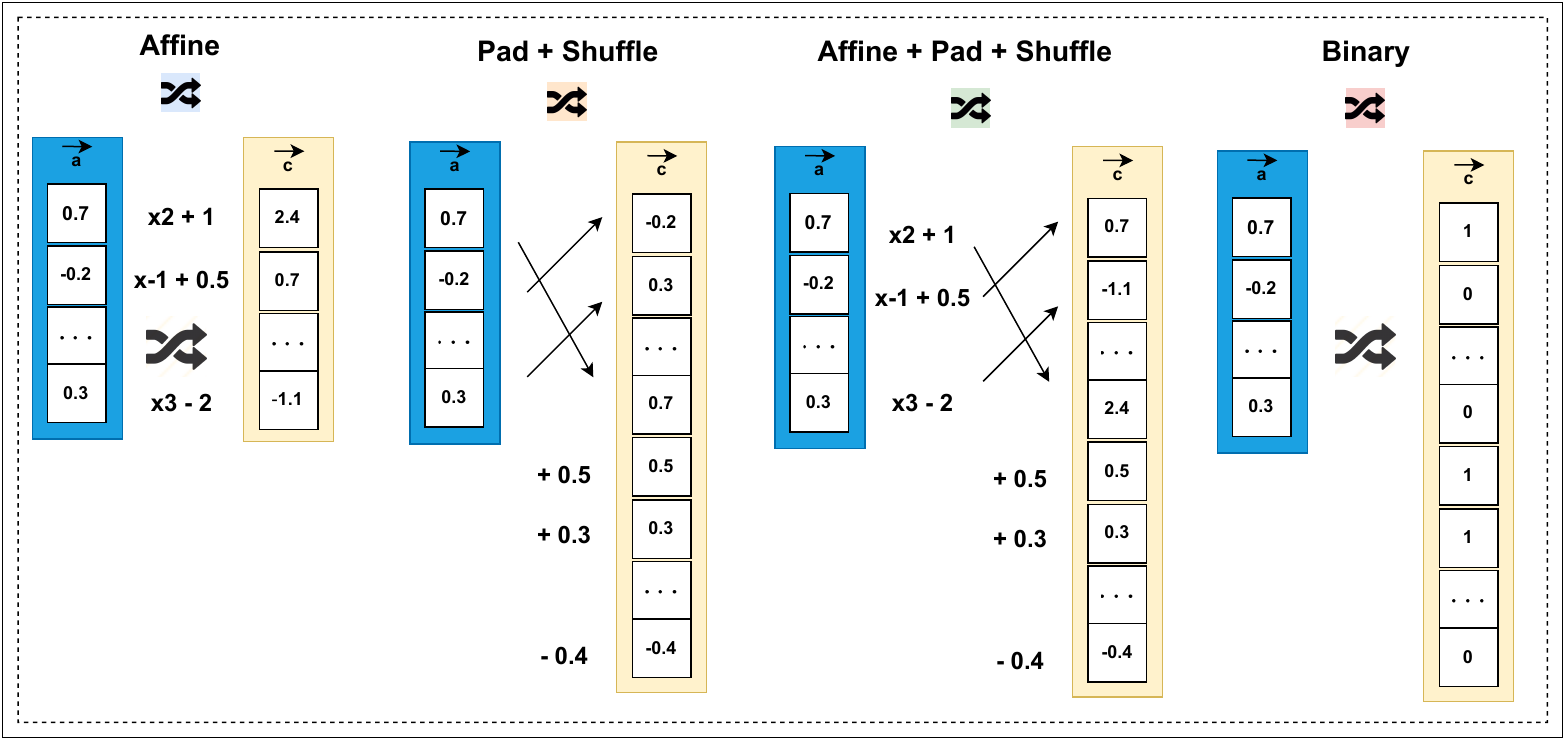} 

    \caption{\textbf{Overview on Transformations.} We depict the inner-workings of the transformations considered in this work.}
    \vspace{-0.in}
    \label{fig:transformations}
\end{figure}

As an alternative to the transformations used within this work (see \Cref{fig:transformations}), one could use a different set of transformations or combinations thereof. The padding can be done with different constant values and combined with adding constant values within the representations. The padding and adding the constant values can be followed by shuffling the elements within the representations. We can apply the affine of binary transformations on top of the padding and shuffling. Additionally, we can also use other pre-defined linear transformations like rotations or shearing. 

The representations could also be compressed to smaller vectors and the compression rate would depend on the occupation of the representation space, for example, the higher the number of occupied buckets in our hash table, the more compressed the output representations could be. Such representations could be compressed via FFT, a cosine transform, or standard compression techniques such as snappy~\citep{snappy}. If the information from the representations should not be lost, then the lossless compression techniques can be applied, for instance, zstd~\citep{zstd}. The only requirement of the compression techniques is to ensure that they do not decrease the accuracy on downstream tasks for legitimate users.

% A supplementary technique that can be integrated with the aforementioned transformations involves masking a small subset of dimensions in the returned representations. These masked dimensions are randomly chosen for each user, which introduces greater complexity to the remapping process as the number of queries employed for remapping increases. However, it is crucial to maintain a low count of masked dimensions within a single representation to ensure its usability for genuine users.

Another alternative is to incorporate an additional neural network layer for transforming the returned representations. The training of this supplementary layer should primarily focus on preserving the usability of the representations for legitimate users. This approach grants the API provider with additional capabilities, as it allows for the utilization of customized training objectives. For instance, if the API provider employs LSH (Locality-Sensitive Hashing) to estimate the coverage of the representation space, they can leverage buckets and train the additional layer to maintain high-quality representations exclusively for frequently-used buckets and their surrounding areas, while not prioritizing the rest of the representation space. This approach safeguards legitimate users from any adverse effects, as their coverage of the representation space is minimal. Simultaneously, it ensures that adversaries are unable to exploit representations from the entire representation space.
% \begin{figure}[h!]
%     \centering 
%     \includegraphics[width=\textwidth, trim={0cm 0cm 0cm 0cm}]{figures/schema/transformations.pdf} 

%     \caption{\textbf{Overview on Transformations.} We depict the inner-workings of the transformations considered in this work.}
%     \vspace{-0.in}
%     \label{fig:transformations}
% \end{figure}
\section{Sybil Attacks}
\label{app:sybils}

We consider an adversary who generates $n$ sybil accounts to steal the encoder from the API. For each of the accounts, the representations are transformed in a different way. Therefore, to replicate the victim model using all the obtained representations, the adversary has to map these representations into one single space. This can be done, for example, by training a neural network to perform the mapping.

We assume the adversary obtains $\{N_1, N_2, \dots, N_n\}$ many representations from the victim for each of the $n$ sybil accounts. Without loss of generality, we assume the adversary maps them back to the embedding space of the first sybil account. To learn the mapping, the adversary can apply different strategies.

\subsection{Sybil Strategies}
We present three potential approaches that Sybils might want to apply to circumvent our defense. Consider three users: $A$, $B$, and $C$, with their respective datasets $D_A$, $D_B$, and $D_C$, each with different distributions to maximize extraction effectiveness. First, user $A$ is selected to unify representations from other users $B$ and $C$. User $A$ would have to query from at least two different datasets $D_B$ and $D_C$, while other users would act legitimately. Sybil attackers want to deploy as many users as possible but with more fake accounts, user $A$ incurs high coverage of the representation space, and this is prevented by our single-user defense. In all other cases neither of the sybil users can act legitimately, thus they are already affected by the single-user defense. Second, user $A$ would query from their own dataset $D_A$ and partially from dataset $D_B$. Then user $B$ would query from their own dataset $D_B$ and partially from dataset $D_C$, and so on. This method is the most inconspicuous but requires a number of remappings that scales super-exponentially with the number of fake accounts, which is impractical.
Finally, each user would query from their respective dataset, for example, user $B$ would query from dataset $D_B$ and additionally from a remapping dataset, \eg $D_A$. Representations could be unified by mapping them to $A$'s representations. 
%The unification of representations could be done, for instance, from each user to representations from user $A$. 
%Third, 
The last approach as well as all other cases reduce to the minimum of remapping between representations of a pair of users. 
%Sybils use many fake user accounts to deploy their attacks, however, 
We show that our defense cuts such attempts short by ensuring that the remapping between representations is prohibitive even for a pair of users.
%In our defense, we cut sybil attempts short and prohibit already the pair-wise remappings.

\section{Additional Related Work}
\label{app:related-work}

One of the main workhorse techniques used in the encoders is contrastive learning, where the representations are trained so that the positive pairs (two augmented versions of the same image) have similar representations while negative pairs (augmentations of two different images) have representations which are far apart. 
%More details description can be found in \Cref{sed:add-related-work}.

\textbf{SimSiam} utilizes Siamese networks (two encoders with shared weights) but with a simplified training process and architecture. In contrast to the previous frameworks, such as SimCLR~\citep{simclr}, SimSiam's authors show that negative samples are unnecessary and collapsing solutions can be avoided by applying the projection head to of one of the encoders, and a stop-gradient operation to the other. 
SimSiam minimizes the negative cosine similarity between two randomly augmented views of the same image from the Siamese encoders, which is expressed via a symmetrized loss~\citep{byol}. This creates a simple yet highly effective representation learning method.

\textbf{DINO} is another popular representation learning framework. While SimSiam uses CNNs, DINO employs vision transformers (ViTs). It trains a student and teacher encoder with the same architecture, updating the teacher with an (exponential moving) average of the student.
Different random transformations of the same image are passed through both encoders. The student receives smaller image crops, forcing it to generate representations restoring parts of the original image. The training objective is minimizing cross-entropy loss between teacher and student representations.

% \textbf{DINO} is another widely used representation learning framework. While SimSiam is trained on Convolutional Neural Networks (CNNs), DINO utilizes vision transformers (ViTs). It trains a student and teacher encoders, both with the same architecture but different parameters, where the teacher is updated with an (exponential moving) average of the student. Different random transformations of the same image are generated and passed through both the student and teacher. The student is provided with smaller crops of the inputs than the teacher, which forces the student to generate representations that restore parts of the initial image. The training objective is to minimize the cross-entropy loss between representations from teacher and student. 

\section{Additional Experimental Results}
\label{app:epxeriments}

\subsection{Details on Experimental Setup}
%\label{app:setup}
%\adam{@Stachu and @Janek: Describe the GPUs used, how many, etc.}
The end-to-end experiments on stealing SimSiam and ViT DINO encoders were done using 3 A100 GPUs. Detailed experiments including mapping, transformations and the evaluation was performed using a single computer equipped with two Nvidia RTX 2080 Ti GPUs. 
%The compute was used for around 6 months.
%Total amount of compute used for these experiments is rather typical for the field.

\subsection{Datasets Used}

\textbf{CIFAR10}~\citep{Krizhevsky09learningmultiple}: The CIFAR10 dataset consists of 32x32 colored images with 10 classes. There are 50000 training images and 10000 test images. 

% \textbf{CIFAR100}~\citep{Krizhevsky09learningmultiple}: The CIFAR100 dataset consists of 32x32 coloured images with 100 classes. There are 50000 training images and 10000 test images.

\textbf{SVHN}~\citep{svhn}: The SVHN dataset contains 32x32 coloured images with 10 classes. There are roughly 73000 training images, 26000 test images and 530000 "extra" images. 

\textbf{ImageNet}\citep{deng2009imagenet}: Larger sized coloured images with 1000 classes. As is commonly done, we resize all images to be of size 224x224. There are approximately 1 million training images and 50000 test images. 

\textbf{STL10}~\citep{coates2011stl10}: The STL10 dataset contains 96x96 coloured images with 10 classes. There are 5000 training images, 8000 test images, and 100000 unlabeled images. 

\textbf{LAION-5B} \cite{schuhmann2022laion5b} The LAION-5B dataset consists of 5,85 billion CLIP-filtered image-text pairs. The dataset was crawled from publically available internet. 

\subsection{More Results for the End2End Empirical Evaluation}

We consider fine-tuning parameters $\beta$, $\lambda$, and $\alpha$ for our cost function and the intuitive meaning behind these parameters. In general, our recommendation is to adjust the parameter $\beta$ that specifies how many buckets are allowed to be filled by users' downstream tasks. 
On the other hand, when parameter $\lambda$ is increased, this causes a higher amount of added noise before we reach the number of buckets specified by $\beta$, which lowers the performance of a given downstream task relatively early. For example, a higher value of $\lambda$ in \Cref{fig:cost-bucket}, would cause an increase in the amount of added noise much earlier than for the target value of $\beta$. Finally, parameter $\alpha$ controls the amount of noise once the number of buckets specified by $\beta$ is reached. Thus, in \Cref{fig:cost-bucket}, we set $\alpha=1$ and the maximum standard deviation of the added Gaussian noise is $1$.

\begin{figure}[h!
]
\vspace{-0.2cm}
        \begin{subfigure}[b]{0.5\textwidth}
            \centering
            \includegraphics[width=1.0\textwidth, trim={0cm 0cm 0cm 0cm}]{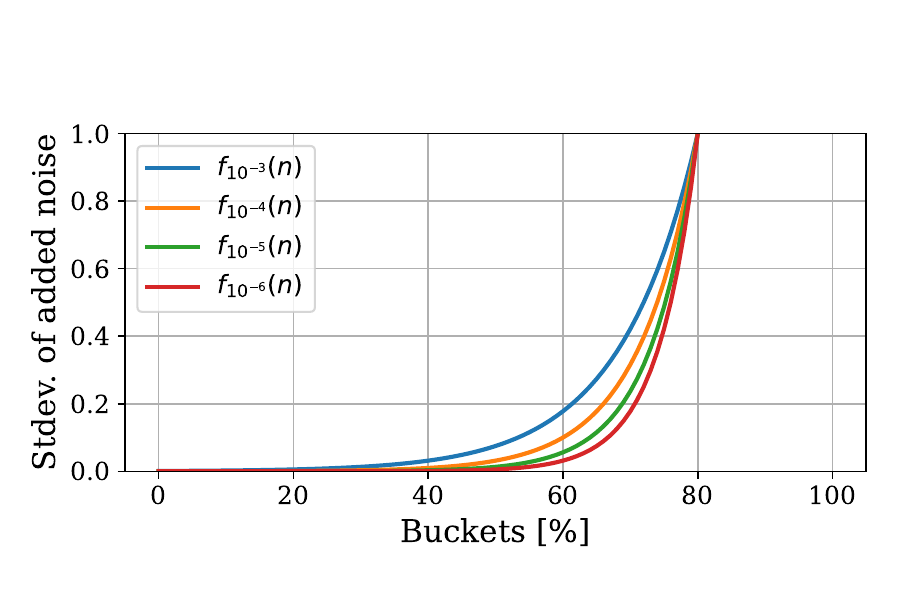}
            \vspace{-0.9cm}
            \caption[]%
            {{\small Cost Function for different $\lambda$ parameter values.}}    
            \label{fig:bucket-lambda}
        \end{subfigure}
        \begin{subfigure}[b]{0.5\textwidth}
            \centering
            \includegraphics[width=1.0\textwidth, trim={0cm 0cm 0cm 0cm}]{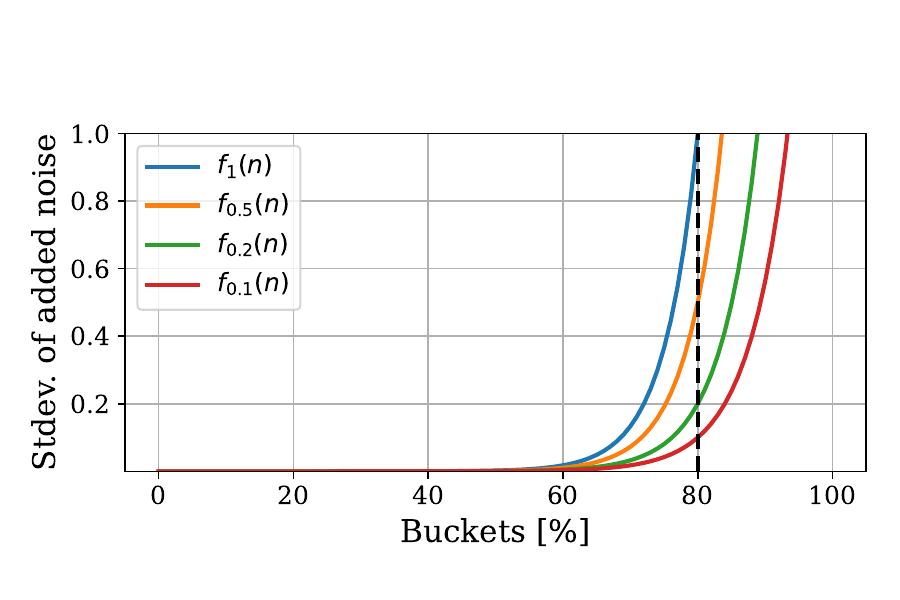}
            \vspace{-0.9cm}
            \caption[]%
            {{\small Cost Function for different $\alpha$ parameter values.}}    
            \label{fig:bucket-histogram}
        \end{subfigure}
    \vspace{-0.1in}
    \caption{\textbf{Effects of $\lambda$ and $\alpha$ parameters on the Cost Function.} 
We present the Cost Function for  $\alpha$=1, $\beta$=80 and different values of $\lambda$ (\textit{left}) and  $\lambda=10^{-6}$, $\beta$=80 and different values of $\alpha$ (\textit{right}).}
    \label{fig:tuning-lambda-and-alfa}
\end{figure}

\definecolor{mygray}{gray}{0.5}
\addtolength{\tabcolsep}{-3pt} 
\begin{table}[hb]
\caption{
\textbf{Stealing and Using Encoders With and Without our Defense}. 
%[Simsiam, $10^-4$, 1, 80]} 
The model used in the experiments is Simsiam, with the following parameters for the cost function $\lambda=10^{-4}$, $\alpha=1$, and $\beta=80$\%, and the number of buckets equal to $2^{12}$.
Due to the higher value of the parameter $\lambda$, we observe lower performance on downstream tasks for the attackers since the magnitude of noise added to the representations is higher.
However, for more complicated tasks than CIFAR10, this change might cause a potential drop in accuracy for the legitimate users.
}
% \caption{\textbf{Stealing and Using Encoders With and Without our Defense}. The model used in the experiments is DINO, with the following parameters for the cost function $\lambda=10^{-6}$, $\alpha=1$, and $\beta=50$\%, and the number of buckets equal to $2^{12}$.
% }
\label{tab:StealImagenet_0.0001_1_80}
\begin{center}
\small
\begin{sc}
\begin{tabu}{ccccccccc}
\toprule
User&Defense& \# Queries & Dataset & Type & CIFAR10 &  STL10 & SVHN & F-MNIST\\
\midrule
\rowfont{\color{mygray}} \Legit & None &  All &  Task & Query &  90.41{\tiny $\pm$0.02} & 95.08{\tiny$\pm$0.13} & 75.47{\tiny $\pm$0.04} & 91.22{\tiny $\pm$0.11} \\
% N/A&N/A& \textit{Victim} & N/A & \textit{ImageNet} & \textit{Train} & 90.49 &  94.9 & 74.98 & 90.57 \\
%\hdashline
%\rowfont{\color{mygray}} \Legit &None &50k & N/A  & CIFAR10  &  & N/A & N/A & N/A\\
\Legit &\name &ALL & TASK & Query & 90.02 {\tiny $\pm$0.1} & 94.88  {\tiny $\pm$0.17} & 74.72  {\tiny $\pm$0.13} & 91.76 {\tiny $\pm$0.09}\\
% \Legit &\name &50k & CIFAR10 & Query & 90.02 {\tiny $\pm$0.1} & N/A & N/A & N/A\\
% %\rowfont{\color{mygray}} \Legit & None &50k & N/A  & STL10  & N/A &  & N/A & N/A\\
% \Legit & \name &5k & STL10 & Query & N/A & 94.88  {\tiny $\pm$0.17} & N/A & N/A\\
% %\rowfont{\color{mygray}} \Legit & None &50k & N/A  & SVHN  & N/A & N/A &  & N/A\\
% \Legit & \name &73k & SVHN & Query & N/A & N/A & 74.72  {\tiny $\pm$0.13}& N/A\\
% %\rowfont{\color{mygray}} \Legit & None &50k & N/A  &  F-MNIST   & N/A & N/A & N/A &\\
% \Legit & \name &60k &  F-MNIST & Query  & N/A & N/A & N/A & 91.76 {\tiny $\pm$0.09}\\
% %\name &50k (single user)& N/A  & STL10   \\
% %\name &60k (single user)& N/A  & F-MNIST   \\
\cdashlinelr{1-9}
\rowfont{\color{mygray}} \Attacker & None & 50K & ImgNet &Steal & 65.2 {\tiny $\pm$0.03} & 64.9 {\tiny $\pm$0.01} & 62.1 {\tiny $\pm$0.01} & 88.5 {\tiny $\pm$0.01}\\
\Attacker &\name &50k & ImgNet &Steal  & 28.22 {\tiny $\pm$ 0.04} & 26.62 {\tiny $\pm$ 0.02} & 19.62 {\tiny $\pm$ 0.02}  & 78.41 {\tiny $\pm$0.01} \\
\rowfont{\color{mygray}} \Attacker & None & 100K & ImgNet &Steal  & 68.1 {\tiny $\pm$0.03} & 63.1 {\tiny $\pm$0.01} & 61.5 {\tiny $\pm$0.01} & 89.0 {\tiny $\pm$0.07}\\
%\cdashlinelr{1-9}
\Attacker &\name &100k & ImgNet &Steal  & 17.73 {\tiny $\pm$ 0.18}  & 15.59 {\tiny $\pm$ 0.61} & 19.53{\tiny $\pm$ 0.01} & 55.11 {\tiny $\pm$ 0.05}\\
% \cdashlinelr{1-8}
% Our Defense (sybil) &25k + 25k& InfoNCE  & ImageNet   & & & & \\
\Sybil &\name &50k+50k & ImgNet &Steal  & 33.43 {\tiny $\pm$ 0.03} & 31.18 {\tiny $\pm$ 0.12} & 22.91 {\tiny $\pm$ 0.01} & 75.35 {\tiny $\pm$ 0.05} \\
\bottomrule
\end{tabu}
\end{sc}
\end{center}
\end{table}
\addtolength{\tabcolsep}{3pt} 

\addtolength{\tabcolsep}{-3pt} 
\begin{table}[ht]
\caption{
\textbf{Stealing and Using Encoders With and Without our Defense}. 
%[Simsiam, $10^-6$, 1, 50]} 
The model used in the experiments is Simsiam, with the following parameters for the cost function $\lambda=10^{-6}$, $\alpha=1$, and $\beta=50$\%, and the number of buckets equal to $2^{12}$.
This experiment corresponds to considering 50\% of buckets filled as a too-large coverage of the embedding space. This improves the defense but again might potentially harm the performance of more complicated tasks than CIFAR10 since they could occupy more buckets than 50\%. 
}
\label{tab:StealImagenet_0.000001_1_50}
\begin{center}
\small
\begin{sc}
\begin{tabu}{ccccccccc}
\toprule
User&Defense& \# Queries & Dataset & Type & CIFAR10 &  STL10 & SVHN & F-MNIST\\
\midrule
\rowfont{\color{mygray}} \Legit & None &  All &  Task & Query  & 90.41{\tiny $\pm$0.02} & 95.08{\tiny$\pm$0.13} & 75.47{\tiny $\pm$0.04} & 91.22 {\tiny $\pm$0.11} \\
% N/A&N/A& \textit{Victim} & N/A & \textit{ImageNet} & \textit{Train} & 90.49 &  94.9 & 74.98 & 90.57 \\
%\cdashlinelr{1-9
%\hdashline
%\rowfont{\color{mygray}} \Legit &None &50k & N/A  & CIFAR10  &  & N/A & N/A & N/A\\
\Legit &\name & ALL& TASK & Query & 90.27 {\tiny $\pm$0.07} & 95.12 {\tiny $\pm$0.13} & 74.94  {\tiny $\pm$0.16}  & 91.66 {\tiny $\pm$0.05}\\
% \Legit &\name &50k & CIFAR10 & Query & 90.27 {\tiny $\pm$0.07} & N/A & N/A & N/A\\
% %\rowfont{\color{mygray}} \Legit & None &50k & N/A  & STL10  & N/A &  & N/A & N/A\\
% \Legit & \name &5k & STL10 & Query & N/A & 95.12 {\tiny $\pm$0.13} & N/A & N/A\\
% %\rowfont{\color{mygray}} \Legit & None &50k & N/A  & SVHN  & N/A & N/A &  & N/A\\
% \Legit & \name &73k & SVHN & Query & N/A & N/A & 74.94  {\tiny $\pm$0.16} & N/A\\
% %\rowfont{\color{mygray}} \Legit & None &50k & N/A  &  F-MNIST   & N/A & N/A & N/A &\\
% \Legit & \name &60k &  F-MNIST & Query  & N/A & N/A & N/A & 91.66 {\tiny $\pm$0.05}\\
% %\name &50k (single user)& N/A  & STL10   \\
% %\name &60k (single user)& N/A  & F-MNIST   \\
\cdashlinelr{1-9}
\rowfont{\color{mygray}} \Attacker & None & 50K & ImgNet &Steal & 65.2 {\tiny $\pm$0.03} & 64.9 {\tiny $\pm$0.01} & 62.1 {\tiny $\pm$0.01} & 88.5 {\tiny $\pm$0.01}\\
\Attacker &\name &50k & ImgNet &Steal  & 15.52 {\tiny $\pm$ 0.37} & 12.57 {\tiny $\pm$ 0.23} & 19.53{\tiny$\pm$ 0.01} & 23.17 {\tiny $\pm$ 0.01}  \\
\rowfont{\color{mygray}} \Attacker & None & 100K & ImgNet &Steal  & 68.1 {\tiny $\pm$0.03} & 63.1 {\tiny $\pm$0.01} & 61.5 {\tiny $\pm$0.01} & 89.0 {\tiny $\pm$0.07}\\
%\cdashlinelr{1-9}
\Attacker &\name &100k & ImgNet &Steal  & 16.27 {\tiny $\pm$ 0.04} & 13.93 {\tiny $\pm$ 0.35} & 19.54 {\tiny $\pm$ 0.02} & 54.69 {\tiny $\pm$ 0.02} \\
% \cdashlinelr{1-8}
% Our Defense (sybil) &25k + 25k& InfoNCE  & ImageNet   & & & & \\
\Sybil &\name &50k+50k & ImgNet &Steal  & 30.14 {\tiny $\pm$ 0.01} & 29.57 {\tiny $\pm$ 0.08} & 19.99 {\tiny $\pm$ 0.03} & 71.72 {\tiny $\pm$ 0.01} \\
\bottomrule
\end{tabu}
\end{sc}
\end{center}
\end{table}
\addtolength{\tabcolsep}{3pt} 

\addtolength{\tabcolsep}{-3pt} 
\begin{table}[ht]

\caption{
\textbf{Stealing and Using Encoders With and Without our Defense}. 
The model used in the experiments is Simsiam, with the following parameters for the cost function $\lambda=10^{-6}$, $\alpha=1$, and $\beta=30$\%, and the number of buckets equal to $2^{12}$.
Since the value of parameter $\beta$ is decreased substantially to 30\%, we observe a drop in accuracy for legitimate users. For example, more than 1\% for CIFAR10. In the next \Cref{tab:StealImagenet_0.000001_0.1_30}, we show that by also decreasing the parameter $\alpha$, we can attenuate this harmful effect and retain higher accuracy for legitimate users. In case of an attack, for 100k stealing queries, we observe much lower accuracy levels than for $\beta=50$\% shown in \Cref{tab:StealImagenet_0.000001_1_50}.
}
\label{tab:StealImagenet_0.000001_1_30}
\begin{center}
\small
\begin{sc}
\begin{tabu}{ccccccccc}
\toprule
User&Defense& \# Queries & Dataset & Type & CIFAR10 &  STL10 & SVHN & F-MNIST\\
\midrule
\rowfont{\color{mygray}} \Legit & None &  All &  Task & Query  & 90.41{\tiny $\pm$0.02} & 95.08{\tiny$\pm$0.13} & 75.47{\tiny $\pm$0.04} & 91.22 {\tiny $\pm$0.11} \\
% N/A&N/A& \textit{Victim} & N/A & \textit{ImageNet} & \textit{Train} & 90.49 &  94.9 & 74.98 & 90.57 \\
%\cdashlinelr{1-9}
%\hdashline
%\rowfont{\color{mygray}} \Legit &None &50k & N/A  & CIFAR10  &  & N/A & N/A & N/A\\
\Legit &\name & ALL & TASK & Query & 88.1 {\tiny $\pm$0.11} & 94.92 {\tiny $\pm$0.11} & 74.37  {\tiny $\pm$0.02} & 91.67 {\tiny $\pm$0.07}\\
% \Legit &\name &50k & CIFAR10 & Query & 88.1 {\tiny $\pm$0.11} & N/A & N/A & N/A\\
% %\rowfont{\color{mygray}} \Legit & None &50k & N/A  & STL10  & N/A &  & N/A & N/A\\
% \Legit & \name &5k & STL10 & Query & N/A & 94.92 {\tiny $\pm$0.11}& N/A & N/A\\
% %\rowfont{\color{mygray}} \Legit & None &50k & N/A  & SVHN  & N/A & N/A &  & N/A\\
% \Legit & \name &73k & SVHN & Query & N/A & N/A & 74.37  {\tiny $\pm$0.02}& N/A\\
% %\rowfont{\color{mygray}} \Legit & None &50k & N/A  &  F-MNIST   & N/A & N/A & N/A &\\
% \Legit & \name &60k &  F-MNIST & Query  & N/A & N/A & N/A & 91.67 {\tiny $\pm$0.07}\\
% %\name &50k (single user)& N/A  & STL10   \\
% %\name &60k (single user)& N/A  & F-MNIST   \\
\cdashlinelr{1-9}
\rowfont{\color{mygray}} \Attacker & None & 50K & ImgNet &Steal & 65.2 {\tiny $\pm$0.03} & 64.9 {\tiny $\pm$0.01} & 62.1 {\tiny $\pm$0.01} & 88.5 {\tiny $\pm$0.01}\\
\Attacker &\name &50k & ImgNet &Steal  & 30.82 {\tiny $\pm$ 0.09} & 26.37 {\tiny $\pm$ 0.07} & 21.87 {\tiny $\pm$ 0.03} & 66.0 {\tiny $\pm$ 0.02} \\
\rowfont{\color{mygray}} \Attacker & None & 100K & ImgNet &Steal  & 68.1 {\tiny $\pm$0.03} & 63.1 {\tiny $\pm$0.01} & 61.5 {\tiny $\pm$0.01} & 89.0 {\tiny $\pm$0.07}\\
%\cdashlinelr{1-9}
\Attacker &\name &100k & ImgNet &Steal  & 9.57 {\tiny $\pm$ 0.17} & 9.83 {\tiny $\pm$ 0.09} & 19.57 {\tiny $\pm$ 0.01} & 27.06 {\tiny $\pm$ 0.46} \\
% \cdashlinelr{1-8}
% Our Defense (sybil) &25k + 25k& InfoNCE  & ImageNet   & & & & \\
\Sybil &\name &50k+50k & ImgNet &Steal  & 29.15 {\tiny $\pm$ 0.02} & 28.67 {\tiny $\pm$ 0.06} & 19.98 {\tiny $\pm$ 0.03} & 70.62 {\tiny $\pm$ 0.03} \\\bottomrule
\end{tabu}
\end{sc}
\end{center}
\end{table}
\addtolength{\tabcolsep}{3pt} 

\clearpage

\addtolength{\tabcolsep}{-3pt} 
\begin{table}[!t]
\vspace{-1cm}
\caption{\textbf{Stealing and Using Encoders With and Without our Defense}. 
%[Simsiam, $10^-6$, 0.1, 30] 
The model used in the experiments is Simsiam, with the following parameters for the cost function $\lambda=10^{-6}$, $\alpha=0.1$, and $\beta=30$\%, and the number of buckets equal to $2^{12}$.
Due to the lower performance on downstream tasks observed in \Cref{tab:StealImagenet_0.000001_1_30} while keeping the parameter $\beta$ fixed to 30\% and $\lambda$ fixed to $10^{-6}$, we decrease the value of parameter $\alpha$ to 0.1, which increases the performance of legitimate users on their downstream tasks.
In this experiment, we also carry out a sybil attack with more accounts (4 instead of 2), but observe that this modification does not improve the performance of the attacker. With more accounts, a sybil has to sacrifice more queries for the remappings between the representations from different accounts. Additionally, note that each account introduces a different remapping error by the dint of different transformations applied to each account by \name.
}
\label{tab:StealImagenet_0.000001_0.1_30}
\begin{center}
\small
\begin{sc}
\begin{tabu}{ccccccccc}
\toprule
User&Defense& \# Queries & Dataset & Type & CIFAR10 &  STL10 & SVHN & F-MNIST\\
\midrule
\rowfont{\color{mygray}} \Legit & None &  All &  Task & Query & 90.41{\tiny $\pm$0.02} & 95.08{\tiny$\pm$0.13} & 75.47{\tiny $\pm$0.04} & 91.22{\tiny $\pm$0.11} \\
% N/A&N/A& \textit{Victim} & N/A & \textit{ImageNet} & \textit{Train} & 90.49 &  94.9 & 74.98 & 90.57 \\
%\cdashlinelr{1-9}
%\hdashline
%\rowfont{\color{mygray}} \Legit &None &50k & N/A  & CIFAR10  &  & N/A & N/A & N/A\\
\Legit &\name &50k & CIFAR10 & Query &  90.17 {\tiny $\pm$0.1} & 94.92  {\tiny $\pm$0.09} & 74.97  {\tiny $\pm$0.13} & 91.71 {\tiny $\pm$0.08}\\
% \Legit &\name &50k & CIFAR10 & Query &  90.17 {\tiny $\pm$0.1} & N/A & N/A & N/A\\
% %\rowfont{\color{mygray}} \Legit & None &50k & N/A  & STL10  & N/A &  & N/A & N/A\\
% \Legit & \name &5k & STL10 & Query & N/A & 94.92  {\tiny $\pm$0.09}& N/A & N/A\\
% %\rowfont{\color{mygray}} \Legit & None &50k & N/A  & SVHN  & N/A & N/A &  & N/A\\
% \Legit & \name &73k & SVHN & Query & N/A & N/A & 74.97  {\tiny $\pm$0.13}& N/A\\
% %\rowfont{\color{mygray}} \Legit & None &50k & N/A  &  F-MNIST   & N/A & N/A & N/A &\\
% \Legit & \name &60k &  F-MNIST & Query  & N/A & N/A & N/A & 91.71 {\tiny $\pm$0.08}\\
% %\name &50k (single user)& N/A  & STL10   \\
% %\name &60k (single user)& N/A  & F-MNIST   \\
\cdashlinelr{1-9}
\rowfont{\color{mygray}} \Attacker & None & 50K & ImgNet &Steal & 65.2 {\tiny $\pm$0.03} & 64.9 {\tiny $\pm$0.01} & 62.1 {\tiny $\pm$0.01} & 88.5 {\tiny $\pm$0.01}\\
\Attacker &\name &50k & ImgNet &Steal  & 19.95 {\tiny $\pm$0.19} & 15.54 {\tiny $\pm$ 0.34} & 19.57 {\tiny $\pm$ 0.01} &  23.50 {\tiny $\pm$ 0.19} \\
\rowfont{\color{mygray}} \Attacker & None & 100K & ImgNet &Steal  & 68.1 {\tiny $\pm$0.03} & 63.1 {\tiny $\pm$0.01} & 61.5 {\tiny $\pm$0.01} & 89.0 {\tiny $\pm$0.07}\\
%\cdashlinelr{1-9}
\Attacker &\name &100k & ImgNet &Steal  & 10.35 {\tiny $\pm$ 0.19} & 12.37 {\tiny $\pm$ 0.69} &19.34 {\tiny $\pm$ 0.01} & 68.93 {\tiny $\pm$ 0.17} \\
% \cdashlinelr{1-8}
% Our Defense (sybil) &25k + 25k& InfoNCE  & ImageNet   & & & & \\
\Sybil &\name &  4$\times$25k & ImgNet &Steal  & 33.15 {\tiny $\pm$ 0.04} & 30.23 {\tiny $\pm$ 0.07} & 20.87 {\tiny $\pm$ 0.01} & 72.19 {\tiny $\pm$ 0.02} \\
\bottomrule
\end{tabu}
\end{sc}
\end{center}
\end{table}
\addtolength{\tabcolsep}{3pt} 

\subsection{\name vs Static Noise Addition Defenses}
\label{app:baseline}

We compare our \name against the current state-of-the-art baseline defense, namely adding a static addition of noise to all the returned representations (as proposed in  \citep{SSLextraction} (Section A.4),\citep{StolenEncoder,ContSteal}). 
For the Table \ref{tab:StealImagenet_noise}, we use the same setup as in Table \ref{tab:StealImagenet} (with an ImageNet pre-trained encoder). 

Our results show the following insights:
\begin{enumerate}
    \item  If the amount of noise is small ($\sigma=0.1$) then the performance drop is negligible but for both a legitimate user (row 2) and an adversary (row 6).  In this case, the defense does not affect the adversary at all (compare rows 5 \& 6).
    \item If the amount of noise is large ($\sigma=10$) then the performance drop is large for both a legitimate user (row 3) and an adversary (row 7). In this case, the encoder is worthless for legitimate users since the performance is too low.
\end{enumerate}

\addtolength{\tabcolsep}{-3pt} 
\begin{table}[hb]
\vspace{0.5cm}
\caption{
\textbf{Stealing and Using Encoders with Static Noise Addition Defenses  vs. Our \name Defense.} 
%[Simsiam, $10^-4$, 1, 80]} 
Adding a small amount of noise results in negligible drop in performance for both legitimate user (row 2) and an adversary (row 6). Adding a large amount of noise defend stealing (row 7), but significantly harm legitimate users at the same time (row 3). Our \name defense solves the above problem and provides high performance for legitimate users (row 4) while effectively defending the encoder against stealing attacks (row 8).
}
\label{tab:StealImagenet_noise}
\begin{center}
\small
\begin{sc}
\begin{tabu}{cccccccccc}
\toprule
User&Defense& \# Queries & Dataset & Type & CIFAR10 &  STL10 & SVHN & F-MNIST\\
\midrule
% \cdashlinelr{1-9}
%\hdashline
%\rowfont{\color{mygray}} \Legit &None &50k & N/A  & CIFAR10  &  & N/A & N/A & N/A\\
\rowfont{\color{mygray}} \Legit & None & ALL & TASK & Query & 90.41 {\tiny $\pm$0.02} & 95.08{\tiny $\pm$0.13} & 75.47{\tiny $\pm$0.04} & 91.22{\tiny $\pm$0.11}\\ 	
%\rowfont{\color{mygray}} \Legit & None &50k & N/A  & STL10  & N/A &  & N/A & N/A\\
 \Legit &  Noise $\sigma$=0.1 & ALL & TASK &  Query & 90.20{\tiny $\pm$0.03}& 95.15{\tiny $\pm$0.13} 	 &75.29{\tiny $\pm$0.09}& 91.24{\tiny $\pm$0.02}\\	 	
%\rowfont{\color{mygray}} \Legit & None &50k & N/A  & SVHN  & N/A & N/A &  & N/A\\
 \Legit &  Noise $\sigma$=10 & ALL & TASK &  Query &65.11{\tiny $\pm$0.45} & 76.37{\tiny $\pm$0.14} &33.23{\tiny $\pm$0.09}& 65.83{\tiny $\pm$0.13} \\ 	 	 	
%\rowfont{\color{mygray}} \Legit & None &50k & N/A  &  F-MNIST   & N/A & N/A & N/A &\\
 \Legit &  \name & ALL & TASK &  Query  & 90.24{\tiny $\pm$0.11}  & 95.05{\tiny $\pm$0.1} & 74.96{\tiny $\pm$0.13} & 91.7{\tiny $\pm$0.01}\\	 	 	
%\name &50k (single user)& N/A  & STL10   \\
%\name &60k (single user)& N/A  & F-MNIST   \\
\cdashlinelr{1-9}
\rowfont{\color{mygray}}  \Attacker &  None & 50K & ImgNet &  Steal & 65.2{\tiny $\pm$0.03} & 64.9{\tiny $\pm$0.01} & 63.1{\tiny $\pm$0.01}  & 88.5 {\tiny $\pm$0.01}\\ 
 \Attacker & Noise $\sigma$=0.1 &50k & ImgNet &  Steal  & 64.92{\tiny $\pm$0.04} & 64.61{\tiny $\pm$0.02} & 62.35{\tiny $\pm$0.01}  & 88.41{\tiny $\pm$0.01} \\ 	 	 	
 \Attacker &  Noise $\sigma$=10 & 50K & ImgNet &  Steal  &36.32{\tiny $\pm$0.2}  &32.59{\tiny $\pm$0.06} & 20.59{\tiny $\pm$0.01}& 74.94{\tiny $\pm$0.02} \\	 	 	
%\cdashlinelr{1-9}
 \Attacker & \name &50k & ImgNet & Steal  & 35.72{\tiny $\pm$0.04}  & 31.54{\tiny $\pm$0.02} & 19.74{\tiny $\pm$0.02}  & 70.01{\tiny $\pm$0.01} \\ 	 		
\bottomrule
\end{tabu}
\end{sc}
\end{center}
\end{table}
\addtolength{\tabcolsep}{3pt} 
\vspace{1.cm}

% are you here?
\subsection{Additional Embedding Space Coverage experiments}
\label{app:coverage-space}

We present additional experiments on measuring the coverage of the representation space. 

\begin{figure}[h!]
    \centering 
    \includegraphics[width=0.5\textwidth, trim={0cm 0.5cm 0cm 1.5cm}, clip]{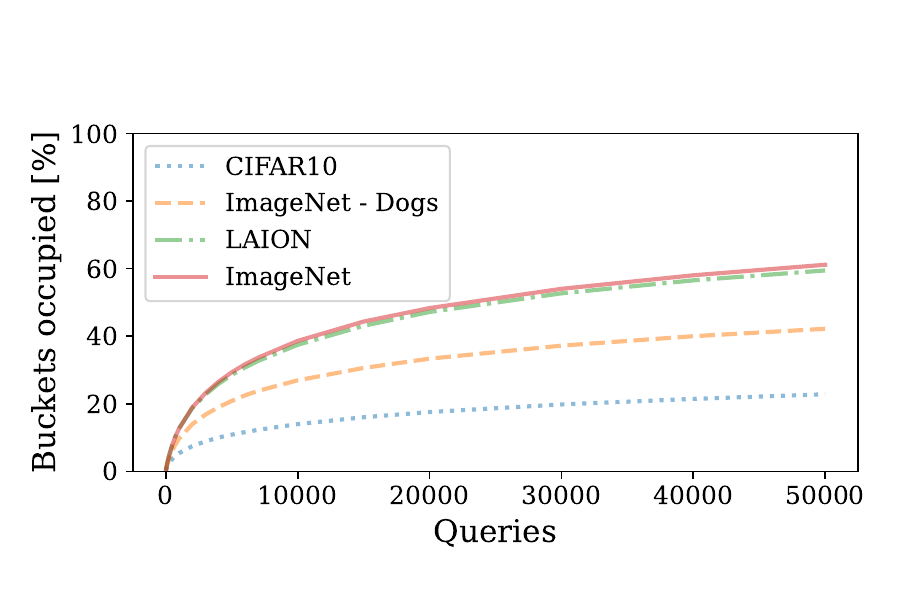} 
    \caption{\textbf{Fraction of Occupied Buckets (Embedding Space Coverage) for the ImageNet encoder.} Representations for the downstream datasets (CIFAR10, ImageNet - Dogs) occupy a smaller fraction of buckets than representations from the complex ImageNet or LAION-5B datasets. The underlying encoder is SimSiam pre-trained on ImageNet with ResNet50.}
    \label{fig:imagenet}
\end{figure}

First, we use the same set-up as from Table \ref{tab:StealImagenet} - SimSiam with ResNet50 pretrained on ImageNet. When querying the encoder with ImageNet-Full (includes all 1000 classes) and LAION-5B datasets, they both occupy a large fraction of the representation space of the victim encoder, as shown on Figure \ref{fig:imagenet}. In contrast, CIFAR10 covers the smallest portion of the representation space as the simplest dataset tested. ImageNet-Dogs (with only 118 classes for dog breeds) falls in the middle, occupying more space than CIFAR10 but less than ImageNet-Full and LAION-5B. Its intermediate coverage aligns with its mid-level difficulty compared to the other datasets. As indicated by representation space coverage, stealing the encoder is similarly effective with ImageNet-Full and LAION-5B datasets, as both datasets cover a large fraction of the representation space. Overall, Figure \ref{fig:imagenet} demonstrates that: 1) our \name can successfully protect the encoder model even from attackers stealing with data that was not used to train the model (LAION-5B in this case) and 2) while providing clean representation for users querying from downstream tasks that are part of more complicated datasets (ImageNet-Dogs).

\vspace{0cm}

\begin{figure}[h!]
    \centering 
    \includegraphics[width=0.5\textwidth, trim={0cm 0.5cm 0cm 1.5cm}, clip]{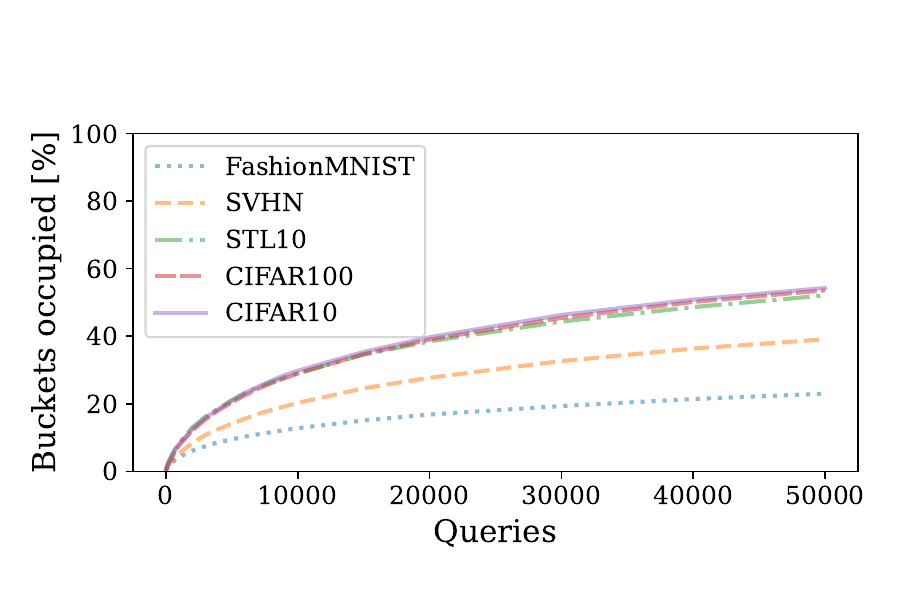} 
    \caption{\textbf{Fraction of Occupied Buckets (Embedding Space Coverage) for the CIFAR10 encoder.} B4B can be applied to an encoder trained on CIFAR10. Representations for the downstream datasets (FashionMNIST, SVHN) occupy a smaller fraction of buckets than representations from CIFAR10, CIFAR100, and STL10 datasets.
    The underlying encoder is SimCLR pre-trained on CIFAR10 with ResNet34.}
    \label{fig:cifar10}
\end{figure}

Our method of measuring the embedding space coverage is not limited to a particular encoder or dataset used for pretraining. We demonstrate this in Figure \ref{fig:cifar10}, showing the fraction of occupied buckets for a SimCLR \cite{chen2020simple} Resnet34 encoder pretrained on CIFAR10.

\clearpage

\subsection{Setting the number of buckets}
\label{app:set-number-of-buckets}
\begin{figure}[!h]
        \centering
        \begin{subfigure}[b]{0.475\textwidth}
      \centering 
    \includegraphics[width=\textwidth, trim={0cm 0.8cm 0cm 1cm}, clip]{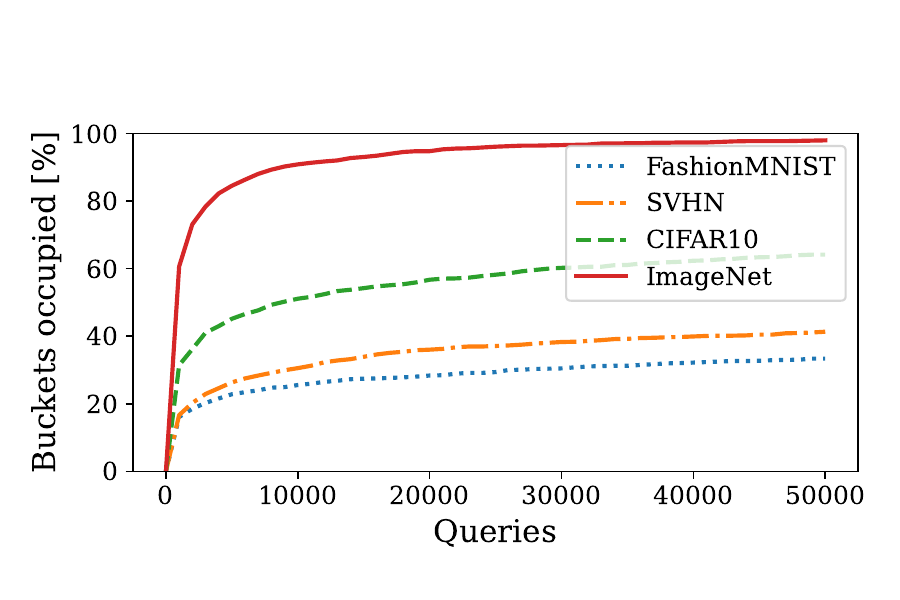} 

    \caption{\textbf{Number of buckets = $2^8$} }
    \vspace{-0.2in}
    \label{fig:buckets_8}
    
        \end{subfigure}
        \hfill
        \begin{subfigure}[b]{0.475\textwidth}  
              \centering 
    \includegraphics[width=\textwidth, trim={0cm 0.8cm 0cm 1cm}, clip]{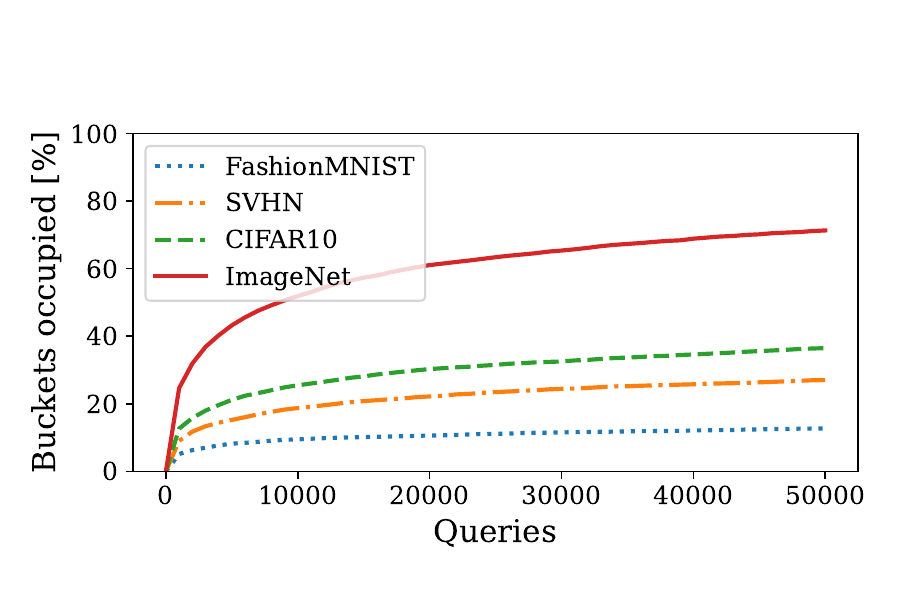} 

    \caption{\textbf{Number of buckets = $2^{10}$} }
    \vspace{-0.2in}
    \label{fig:buckets_10}
        \end{subfigure}
        \vskip\baselineskip
        \begin{subfigure}[b]{0.475\textwidth}   
              \centering 
    \includegraphics[width=\textwidth, trim={0cm 0.8cm 0cm 1cm}, clip]{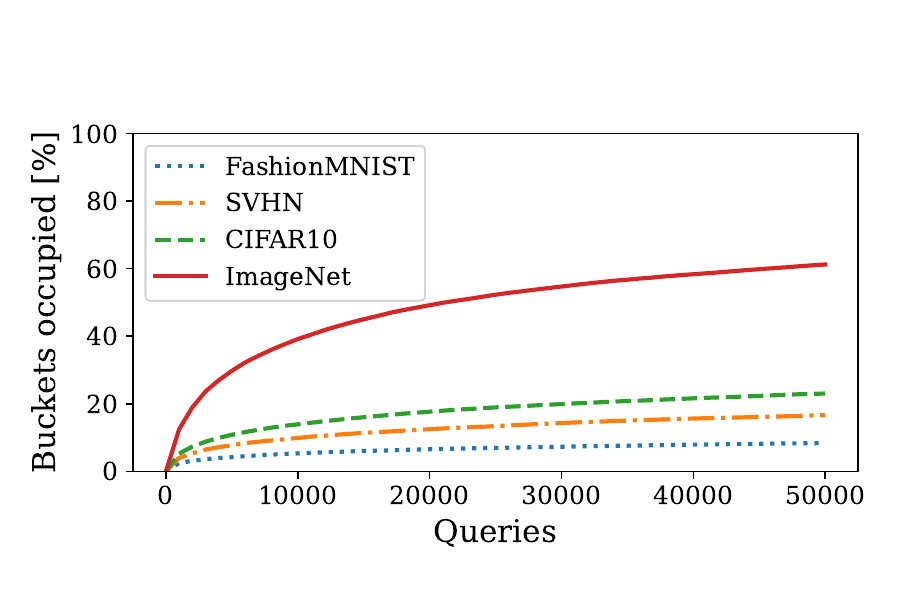} 

    \caption{\textbf{Number of buckets = $2^{12}$} }
    \vspace{-0in}
    \label{fig:buckets_12}
        \end{subfigure}
        \hfill
        \begin{subfigure}[b]{0.475\textwidth}   
               \centering 
    \includegraphics[width=\textwidth, trim={0cm 0.8cm 0cm 1cm}, clip]{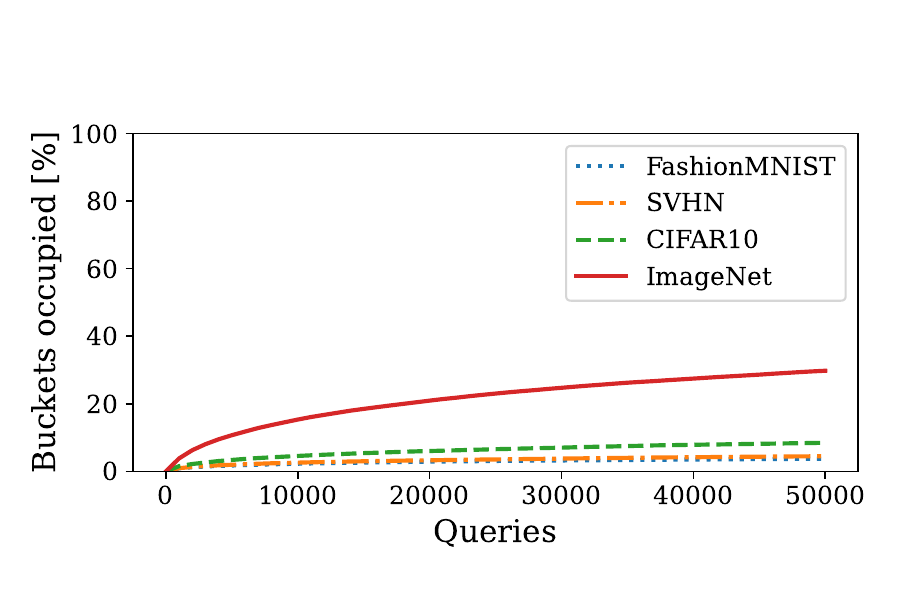} 

    \caption{\textbf{Number of buckets = $2^{14}$} }
    \vspace{-0in}
    \label{fig:buckets_14}
        \end{subfigure}
        \caption{
        \textbf{Estimating Embedding Space Coverage through LSH on the SimSiam Encoder.} We extend the results from \Cref{fig:bucket-query-number}(a) and present the fraction of buckets occupied by representations of different datasets as a function of the number of queries posed to the encoder. We consider different number of buckets in the LSH table. We observe that $2^8$ buckets is to small since queries from the ImageNet dataset saturate all the buckets after around 50k queries, while the number $2^{14}$ of buckets is too large since it is never occupied more than 40\%. Thus, the number $2^{12}$ buckets is a good middle ground.
        Subfigure (c) corresponds to \Cref{fig:bucket-query-number} from the main paper.
        We also use the same notation and carry out our experiments in the same way as in \Cref{fig:bucket-query-number}.
        %\janek{caption}
        } 
        \label{fig:number of total buckets}
    \end{figure}

We present our procedure to find an optimal number of buckets in \Cref{fig:number of total buckets}.

% \begin{figure}[h!]

%     \centering 
%     \includegraphics[width=0.4\textwidth, trim={0cm 0cm 0cm 0cm}]{figures/number_of_buckets/buckets_occupied_simsiam-formating_8.pdf} 

%     \caption{\textbf{n=8} }
%     \vspace{-0.in}
%     \label{fig:buckets_8}
% \end{figure}

% \begin{figure}[h!]
%     \centering 
%     \includegraphics[width=0.4\textwidth, trim={0cm 0cm 0cm 0cm}]{figures/number_of_buckets/buckets_occupied_simsiam-formating_10.pdf} 

%     \caption{\textbf{n=10} }
%     \vspace{-0.in}
%     \label{fig:buckets_10}
% \end{figure}

% \begin{figure}[h!]
%     \centering 
%     \includegraphics[width=0.4\textwidth, trim={0cm 0cm 0cm 0cm}]{figures/number_of_buckets/buckets_occupied_simsiam-formating_13.pdf} 

%     \caption{\textbf{n=13 - main paper} }
%     \vspace{-0.in}
%     \label{fig:buckets_13}
% \end{figure}

% \begin{figure}[h!]
%     \centering 
%     \includegraphics[width=0.4\textwidth, trim={0cm 0cm 0cm 0cm}]{figures/number_of_buckets/buckets_occupied_simsiam-formating_14.pdf} 

%     \caption{\textbf{n=14} }
%     \vspace{-0.in}
%     \label{fig:buckets_14}
% \end{figure}

% \begin{figure}[h!]
%     \centering 
%     \includegraphics[width=0.4\textwidth, trim={0cm 0cm 0cm 0cm}]{figures/number_of_buckets/buckets_occupied_simsiam-formating_15.pdf} 

%     \caption{\textbf{n=15} }
%     \vspace{-0.in}
%     \label{fig:buckets_15}
% \end{figure}

\subsection{Results for DINO}

\begin{figure}[!h]
    \vspace{-0.5cm}
    \centering 
    \includegraphics[width=0.5\textwidth, trim={0cm 0cm 0cm 0cm}, clip]{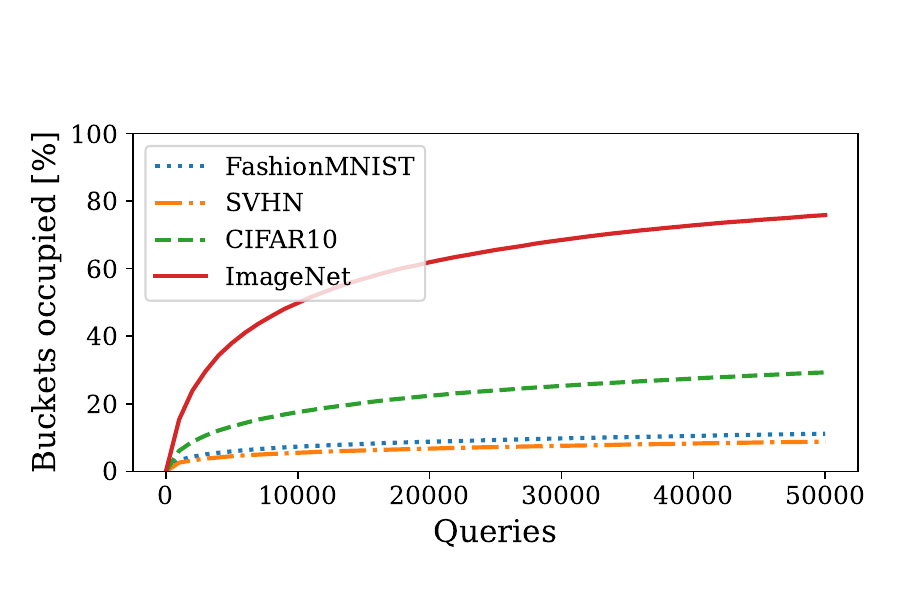} 

    \caption{\textbf{Estimating Embedding Space Coverage through LSH on the DINO Encoder.} 
    The number of buckets is set to $2^{12}$.
    We also use the same notation and carry out our experiments in the same way as in \Cref{fig:bucket-query-number}.
    }
    \vspace{-0in}
    \label{fig:buckets_dino}
\end{figure}

We show that our defense is also applicable to the DINO encoder. The occupation of the representations space is presented visually in \Cref{fig:clusters-dino}. We also show that the number of buckets $2^{12}$ is optimal for DINO in \Cref{fig:buckets_dino}. 
The impact of transformation on the representations from DINO is shown \Cref{tab:LegitimateTransformations}.
% in \Cref{fig:transformations_debinarization} and \Cref{fig:transformations_debinarization-cos-dist}.
Finally, the end to end experiment for DINO is presented in \Cref{tab:StealImagenet_DINO_0.000001_1_50}.

\definecolor{mygreen2}{RGB}{0, 180, 18}
\begin{figure}[h!]%{r}{4cm}
%\begin{figure}[h!]
\vspace{0cm}
    \centering
    \includegraphics[width=0.25\textwidth, trim={0cm 0cm 0cm 0cm}]{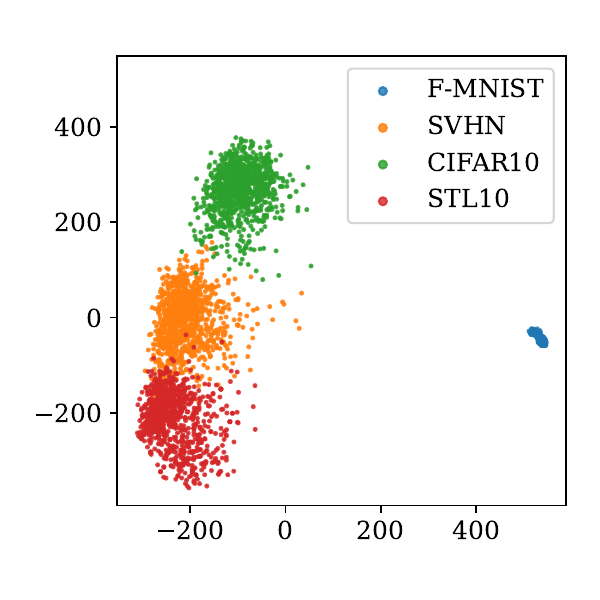} 
    %\vspace{-2ex}
    %\caption{\textbf{Lower Dimension Representation of Different Representations.} We map the representations obtained for different downstream tasks to a two-dimensional space. We observe that the different downstream tasks form clusters.}
    \caption{\textbf{Representations from Different Tasks Occupy Different Sub-Spaces of the Embedding Space. Presented for {\color{blue}FashionMNIST}, {\color{orange}SVHN}, {\color{mygreen2}CIFAR10}, and {\color{red}STL10}.} In this plot, we used the DINO ViT Small encoder trained on ImageNet.
    %\janek{Please choose pca from full representation (encoder hidden state) - bottom, or last 4 layers - top}
    %\adam{The legend has to be changed. We should have simply: FashionMNIST, STL10, SVHN, and CIFAR10. If we add ImageNet, it would probably cover more of the representation space - this might be shown in a separate picture.}
    %\franzi{I need to color the names}
    %\ta{Why do we have this figure instead of the one with the very small space occupied for FashionMNIST?}
    }
    \vspace{-0.2cm}
    \label{fig:clusters-dino}
%\end{figure}
\end{figure}

\addtolength{\tabcolsep}{-3pt} 
\begin{table}[h!]
\caption{
\textbf{Stealing and Using Encoders With and Without our Defense}. 
The model used in the experiments is DINO, with the following parameters for the cost function $\lambda=10^{-6}$, $\alpha=1000$, and $\beta=60$\%, and the number of buckets equal to $2^{12}$.
We have to increase the value of parameter $\alpha$ by $\times$1000 since the norms of the DINO representations are also around $10^{3}$ higher than for SimSiam.
We observe that \name performs similarly on DINO as for SimSiam.
%(in this case, for the stricter settings of our defense).
}
\label{tab:StealImagenet_DINO_0.000001_1_50}
\begin{center}
\small
\begin{sc}
\begin{tabu}{ccccccccc}
\toprule
User&Defense& \# Queries & Dataset & Type & CIFAR10 &  STL10 & SVHN & F-MNIST\\
\midrule
\rowfont{\color{mygray}} \Legit & None &  All &  Task & Query & 94.51 {\tiny $\pm$0.08} & 97.98 {\tiny $\pm$0.04} & 70.66 {\tiny $\pm$0.16} & 89.98 {\tiny $\pm$0.03} \\
% N/A&N/A& \textit{Victim} & N/A & \textit{ImageNet} & \textit{Train} & 90.49 &  94.9 & 74.98 & 90.57 \\
%\cdashlinelr{1-9}
%\hdashline
\Legit &\name & ALL & TASK & Query & 94.25 {\tiny $\pm$0.11} &  98.05 {\tiny $\pm$0.04} & 69.66 {\tiny $\pm$0.14} & 89.68 {\tiny $\pm$0.01}\\
% %\rowfont{\color{mygray}} \Legit &None &50k & N/A  & CIFAR10  &  & N/A & N/A & N/A\\
% \Legit &\name &50k & CIFAR10 & Query & 94.25 {\tiny $\pm$0.11} & N/A & N/A & N/A\\
% %\rowfont{\color{mygray}} \Legit & None &50k & N/A  & STL10  & N/A &  & N/A & N/A\\
% \Legit & \name &5k & STL10 & Query & N/A & 98.05 {\tiny $\pm$0.04} & N/A & N/A\\
% %\rowfont{\color{mygray}} \Legit & None &50k & N/A  & SVHN  & N/A & N/A &  & N/A\\
% \Legit & \name &73k & SVHN & Query & N/A & N/A & 69.66 {\tiny $\pm$0.14}& N/A\\
% %\rowfont{\color{mygray}} \Legit & None &50k & N/A  &  F-MNIST   & N/A & N/A & N/A &\\
% \Legit & \name &60k &  F-MNIST & Query  & N/A & N/A & N/A & 89.68 {\tiny $\pm$0.01}\\
%\name &50k (single user)& N/A  & STL10   \\
%\name &60k (single user)& N/A  & F-MNIST   \\
\cdashlinelr{1-9}
\rowfont{\color{mygray}} \Attacker & None & 50K & ImgNet &Steal & 67.92 {\tiny $\pm$ 0.04} & 66.02 {\tiny $\pm$ 0.22} & 61.30 {\tiny $\pm$ 0.01} &  89.46 {\tiny $\pm$ 0.01} \\
\Attacker &\name &50k & ImgNet &Steal   & 42.02{\tiny$\pm$0.05} & 38.91{\tiny$\pm$0.06} & 19.94{\tiny$\pm$0.02} & 73.33{\tiny$\pm$0.04} \\
\rowfont{\color{mygray}} \Attacker & None & 100K & ImgNet &Steal  & 75.07 {\tiny $\pm$ 0.01} & 76.32 {\tiny $\pm$ 0.02} & 71.79 {\tiny $\pm$ 0.06}  & 89.76 {\tiny $\pm$ 0.01}\\
%\cdashlinelr{1-9}
\Attacker &\name &100k & ImgNet &Steal    & 19.27{\tiny$\pm$0.03} & 21.24{\tiny $\pm$0.03} & 19.84\tiny{$\pm$0.01} & 71.01{\tiny$\pm$0.03} \\
% \cdashlinelr{1-8}
% Our Defense (sybil) &25k + 25k& InfoNCE  & ImageNet   & & & & \\
\Sybil &\name & 50k+50k & ImgNet &Steal  & 45.56{\tiny$\pm$ 0.06} & 42.50{\tiny$\pm$0.02} & 24.25{\tiny$\pm$0.03} & 78.01{\tiny$\pm$ 0.08} \\
\bottomrule
\end{tabu}
\end{sc}
\end{center}
\end{table}
\addtolength{\tabcolsep}{3pt} 

% \begin{figure}[h!]
%     \centering 
%     \includegraphics[width=0.4\textwidth, trim={0cm 0cm 0cm 0cm}]{figures/dino/pca_dataset_clusters_stl10_dino.pdf} 

%     \caption{\textbf{Cluster for DINO. Embedding dim 1536} }
%     \vspace{-0.in}
%     \label{fig:buckets_dino}
% \end{figure}

% \begin{figure}[h!]
%     \centering 
%     \includegraphics[width=0.4\textwidth, trim={0cm 0cm 0cm 0cm}]{figures/clusters/full_repr_dataset_clusters_stl10_4x4.pdf} 

%     \caption{\textbf{Cluster for DINO. Embedding dim Full hidden state} }
%     \vspace{-0.in}
%     \label{fig:buckets_dino}
% \end{figure}

\subsection{Additional evaluation of transformations}

Additionally, we show the impact of transformations on the performance of legitimate users in \Cref{tab:LegitimateTransformations} (for both SimSiam and DINO).

\begin{figure}[bh]
    \centering
    \includegraphics[width=\textwidth, trim={0cm 0cm 0cm 0cm}]{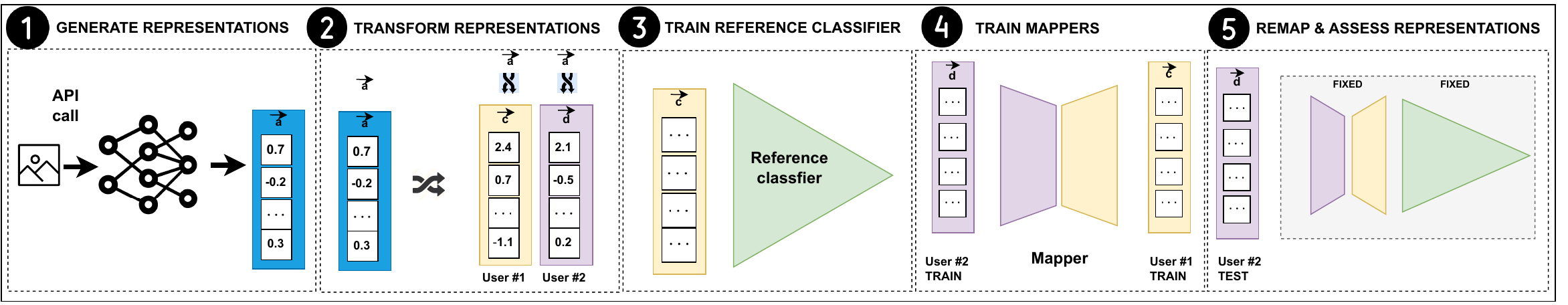} 

    \caption{\textbf{Protocol to Evaluate the Mapping Between Representations.} We present the protocol of evaluating remappings for two sybil accounts. \encircle{1} API receives inputs from two sybil accounts and generates corresponding representations. \encircle{2} Representations are transformed on a per-user basis and returned. \encircle{3} Adversary trains a reference classifier on representations from account one. \encircle{4} Adversary trains a linear model to find mapping from representations of account two to representations of account one. \encircle{5} To check the quality of obtained mapping  representations from test set of account two are mapped using the fixed mapper (from step 4) to representation space of account one. This enables the calculation of cosine distance between representations from account one and their counterparts from account two shown in Figure \ref{fig:quality-remap-main}. Additionally, the fixed reference classifier (from step 3) can be used to measure the accuracy drop caused by remapping . 
    }
    %\stachu{This is not the case. Normally, attacker would probably do sth like this. Here, in this setup we only want to evaluate how good the mapping could be given N queries or in other words, how much (acc) do we lose on mapping in comparison to normal user acc.
    % The goal is to show how much additional queries attacker needs to have a good mapping between accounts.
    % We do not train a classifier or a mapper in step 5 - they are fixed (from step 3 and 4). If we have drop in acc for account two in point 5 compared to acc obtained by account one on trained reference model, it shows that our mapping is not ideal. So the plots in main part shows the effort attacker must add - number of additional queries for remapping to ensure there is no drop from step 3 representations acc to step 5 representations acc.}
    %Re-map the transformed representations to the user \#1 space and asses their quality through the proxy of the accuracy obtained on the reference classifier trained in step 3.

    \vspace{-0.in}
    \label{fig:mapping_setup}
\end{figure}

\makeatletter
\setlength{\@fptop}{0pt}
\makeatother

\definecolor{mygray}{gray}{0.5}
\addtolength{\tabcolsep}{-3.5pt} 
\begin{table}[ht!]
\caption{\textbf{Impact of Transformations on the Performance for Legitimate Users.} 
We show that the transformations applied per-account do not harm the performance of legitimate users on their downstream tasks.
The victim encoders was trained on the ImageNet dataset using SimSiam and DINO frameworks.
%\adam{We can also add the results for DINO, then we have to include one more column with the name of the encoder: SimSiam or DINO.}
}
\label{tab:LegitimateTransformations}
\begin{center}
\begin{small}
\begin{sc}
\begin{tabu}{cccccc}
\toprule
Transformation & Encoder & CIFAR10 & STL10 & SVHN & F-MNIST\\
\midrule
% N/A&N/A& \textit{Victim} & N/A & \textit{ImageNet} & \textit{Train} & 90.33 &  94.9 & 79.39 & 91.9 \\
None & \textit{Victim SimSiam}  & 90.41{\tiny $\pm$0.02} & 95.08 {\tiny$\pm$0.13} & 75.47 {\tiny $\pm$0.04} & 91.22 {\tiny $\pm$0.11}\\
\cdashlinelr{1-6}
%\rowfont{\color{mygray}} \Legit &None &50k & N/A  & CIFAR10  &  & N/A & N/A & N/A\\
Affine & SimSiam & 90.24 {\tiny $\pm$0.11} & 95.05 {\tiny $\pm$0.1} & 74.96 {\tiny $\pm$0.18} & 91.42 {\tiny $\pm$0.15} \\
Pad+Shuffle & SimSiam & 90.4 {\tiny $\pm$0.05} & 95.34 {\tiny $\pm$0.06} & 75.47 {\tiny $\pm$0.01} & 91.38 {\tiny $\pm$0.15}\\
%\rowfont{\color{mygray}} \Legit & None &50k & N/A  & SVHN  & N/A & N/A &  & N/A\\
Affine+Pad+Shuffle & SimSiam & 90.18 {\tiny $\pm$0.06} & 95.03 {\tiny $\pm$0.05} & 74.86 {\tiny $\pm$0.1} & 91.35 {\tiny $\pm$0.1} \\
%\rowfont{\color{mygray}} \Legit & None &50k & N/A  &  F-MNIST   & N/A & N/A & N/A &\\
Binary & SimSiam&  88.78 {\tiny $\pm$0.2} & 94.72 {\tiny $\pm$0.02} & 68.42 {\tiny $\pm$0.16} & 88.91 {\tiny $\pm$0.34}\\
\midrule
None & \textit{Victim DINO}  & 94.51 {\tiny $\pm$0.08} & 97.98 {\tiny $\pm$0.04} & 70.66 {\tiny $\pm$0.16} & 89.98 {\tiny $\pm$0.03}\\
\cdashlinelr{1-6}
%\rowfont{\color{mygray}} \Legit &None &50k & N/A  & CIFAR10  &  & N/A & N/A & N/A\\
Affine & DINO & 94.25 {\tiny $\pm$0.11} & 98.05 {\tiny $\pm$0.04} & 69.77 {\tiny $\pm$0.11} &  89.68 {\tiny $\pm$0.01}\\
Pad+Shuffle & DINO & 94.72 {\tiny $\pm$0.02} & 98.07 {\tiny $\pm$0.03} & 70.44 {\tiny $\pm$0.1} & 89.91 {\tiny $\pm$0.08}\\
%\rowfont{\color{mygray}} \Legit & None &50k & N/A  & SVHN  & N/A & N/A &  & N/A\\
Affine+Pad+Shuffle & DINO & 94.26 {\tiny $\pm$0.06} & 98.02 {\tiny $\pm$0.01} & 69.49 {\tiny $\pm$0.2} &  89.70 {\tiny $\pm$0.1}\\
%\rowfont{\color{mygray}} \Legit & None &50k & N/A  &  F-MNIST   & N/A & N/A & N/A &\\
Binary & DINO &  92.96 {\tiny $\pm$0.1} & 98.03 {\tiny $\pm$0.03} & 59.53 {\tiny $\pm$0.27} & 88.26 {\tiny $\pm$0.04}\\
%\name &50k (single user)& N/A  & STL10   \\
% %\name &60k (single user)& N/A  & F-MNIST   \\
% \cdashlinelr{1-10}
% \rowfont{\color{mygray}} \Attacker &None & 50K & InfoNCE  & ImgNet &Steal & 65.2 & 64.9 & 62.1 & 88.5 \\
% \Attacker &\name &50k& InfoNCE  & ImgNet &Steal  & 15.2 & 14.1 & 19.6 & 25.3 \\
% \rowfont{\color{mygray}} \Attacker &None & 100K & InfoNCE& ImgNet &Steal  & 68.1  & 63.1 & 61.5 & 89.0 \\
% %\cdashlinelr{1-9}
% \Attacker &\name &100k & InfoNCE  & ImgNet &Steal  & 11.6 & 10.3 & 19.6 & 11.6 \\
% % \cdashlinelr{1-8}
% % Our Defense (sybil) &25k + 25k& InfoNCE  & ImageNet   & & & & \\
% \Sybil &\name &50k+50k & InfoNCE  & ImgNet &Steal  & 31.8 & 27.4 & 19.8 & 64.7\\
\bottomrule
\end{tabu}
\end{sc}
\end{small}
\end{center}
\end{table}
\addtolength{\tabcolsep}{3.5pt}

% \ts{Check appendix letters in main paper and unify here}
% We show how the representations from different tasks occupy different sub-spaces of the embedding space for the DINO encoder in \Cref{fig:clusters-dino}.

%\input{contents/appendix-backup}

%\bibliographystyle{plainnat}
%\bibliography{main}

% \bibliographystyle{plainnat}
% \bibliography{main}

\end{document}